\documentclass[sn-mathphys-num]{sn-jnl-seniority}% Math and 
\usepackage{graphicx}%
\usepackage{multirow}%
\usepackage{amsmath,amssymb,amsfonts}%
\usepackage{amsthm}%
\usepackage{mathrsfs}%
\usepackage[title]{appendix}%
\usepackage{xcolor}%
\usepackage{textcomp}%
\usepackage{manyfoot}%
\usepackage{booktabs}%
\usepackage{algorithm}%
\usepackage{algorithmicx}%
\usepackage{algpseudocode}%
\usepackage{listings}%
\usepackage{tikz}
\usepackage{ulem}
\usepackage{epstopdf}
\usepackage{float}

\usepackage{outlines}
\usepackage{soul}
\usepackage{booktabs}
\usepackage{multirow}

\definecolor{myblue}{rgb}{0.36, 0.54, 0.66}

\newcommand{\cmark}{\checkmark}
\usepackage{xspace}
\newcommand{\added}[1]{\textcolor{black}{#1}}

\newcommand{\epic}{Epic-tent\xspace} 
\newcommand{\epicO}{Epic-tent-O\xspace} 
\newcommand{\assembly}{Assembly101\xspace} 
\newcommand{\assemblyO}{Assembly101-O\xspace} 

\DeclareRobustCommand{\myparagraph}[1]{\noindent\textbf{#1}} 
\usepackage[framemethod=TikZ]{mdframed}
\definecolor{Gray}{gray}{0.9}
\definecolor{LightGray}{gray}{0.95}

\newmdenv[
  backgroundcolor=LightGray,
  linewidth=1pt,
  linecolor=Gray,
  roundcorner=7pt
]{myframe}

\definecolor{systemcolor}{RGB}{220, 220, 220}  % Light gray
\definecolor{acotcolor}{RGB}{255, 230, 204}    % Light orange
\definecolor{contextcolor}{RGB}{204, 255, 204} % Light green
\definecolor{sequencecolor}{RGB}{204, 229, 255} % Light blue
\definecolor{bordercolor}{RGB}{150, 150, 150}  % Gray border

\newmdenv[
  backgroundcolor=systemcolor,
  linewidth=1pt,
  linecolor=bordercolor,
  roundcorner=7pt
]{systemframe}

\newmdenv[
  backgroundcolor=acotcolor,
  linewidth=1pt,
  linecolor=bordercolor,
  roundcorner=7pt
]{acotframe}

\newmdenv[
  backgroundcolor=contextcolor,
  linewidth=1pt,
  linecolor=bordercolor,
  roundcorner=7pt
]{contextframe}

\newmdenv[
  backgroundcolor=sequencecolor,
  linewidth=1pt,
  linecolor=bordercolor,
  roundcorner=7pt
]{sequenceframe}

\theoremstyle{thmstyleone}%
\theoremstyle{thmstyletwo}%

\theoremstyle{thmstylethree}%

\begin{document}

\title[TI-PREGO: Chain of Thought and In-Context Learning for Online Mistake Detection in PRocedural EGOcentric Videos]{\textbf{TI-PREGO}: Chain of \textbf{T}hought and \textbf{I}n-Context Learning for Online Mistake Detection in \textbf{PR}ocedural \textbf{EGO}centric Videos}

\author[1,4]{\fnm{Leonardo} \sur{Plini}}\email{leonardo.plini@uniroma1.it}
\equalcont{These authors contributed equally to this work.}

\author[1]{\fnm{Luca} \sur{Scofano}}\email{scofano@diag.uniroma1.it}
\equalcont{These authors contributed equally to this work.}

\author[1]{\fnm{Edoardo} \sur{De Matteis}}\email{dematteis@di.uniroma1.it}
\equalcont{These authors contributed equally to this work.}

\author[1]{\fnm{Guido Maria} \sur{D'Amely di Melendugno}}\email{damely@di.uniroma1.it}

\author[1,2]{\fnm{Alessandro} \sur{Flaborea}}\email{flaborea@di.uniroma1.it}

\author[1]{\fnm{Andrea} \sur{Sanchietti}}\email{sanchietti.1883210@studenti.uniroma1.it}

\author[3]{\fnm{Giovanni Maria} \sur{Farinella}}\email{giovanni.farinella@unict.it}\equalsenior{Co-senior role.}

\author[1]{\fnm{Fabio} \sur{Galasso}}\email{galasso@di.uniroma1.it}\equalsenior{Co-senior role.}

\author[3]{\fnm{Antonino} \sur{Furnari}}\email{antonino.furnari@unict.it}\equalsenior{Co-senior role.}

\affil[1]{\orgdiv{Sapienza University of Rome, Italy}}

\affil[2]{\orgdiv{ItalAI (italailabs.com)}}

\affil[3]{\orgdiv{University of Catania, Italy}}

\affil[4]{\orgdiv{National Institute of Nuclear Physics - LNF, Italy}}

\abstract{
Identifying procedural errors online from egocentric videos is a critical yet challenging task across various domains, including manufacturing, healthcare and skill-based training. The nature of such mistakes is inherently open-set, as unforeseen or novel errors may occur, necessitating robust detection systems that do not rely on prior examples of failure. 
Currently, no existing technique can reliably detect open-set procedural mistakes in an online setting.
We propose a dual-branch architecture to address this problem in an online fashion:
the recognition branch takes input frames from egocentric video, predicts the current action and aggregates frame-level results into action tokens while the anticipation branch leverages the solid pattern-matching capabilities of Large Language Models (LLMs) to predict action tokens based on previously predicted ones.
Mistakes are detected as mismatches between the currently recognized action and the action predicted by the anticipation module.

Extensive experiments on two novel procedural datasets demonstrate the challenges and opportunities of leveraging a dual-branch architecture for mistake detection, showcasing the effectiveness of our proposed approach.
}


\maketitle

% INTRODUCTION
\section{Introduction}\label{sec:intro}

Detecting procedural errors in videos has recently gained significant attention as it holds the potential to enhance safety and efficiency in several fields.
Humans engage in procedural activities everyday in different contexts, such us daily activities (e.g., cooking a meal), hobbies (e.g., fixing a bike), and professional settings (e.g., operating complex machineries). Monitoring procedural activities is crucial to guarantee the desired results and operate safely, especially in high stakes contexts such as industrial workflows. Wearable devices, looking at the scene from the user's point of view, and endowed with the capability of providing feedback through augmented reality, can be used to assess procedural mistakes (e.g., forgetting to add water to the kettle before turning it on) and provide real-time feedback.
This immediate feedback allows for timely corrections, fostering faster skill development, skill acquisition and safer learning environments in high-stakes areas like surgery and industrial workflow.

\begin{figure}
    \centering
    \includegraphics[width=0.8\linewidth]{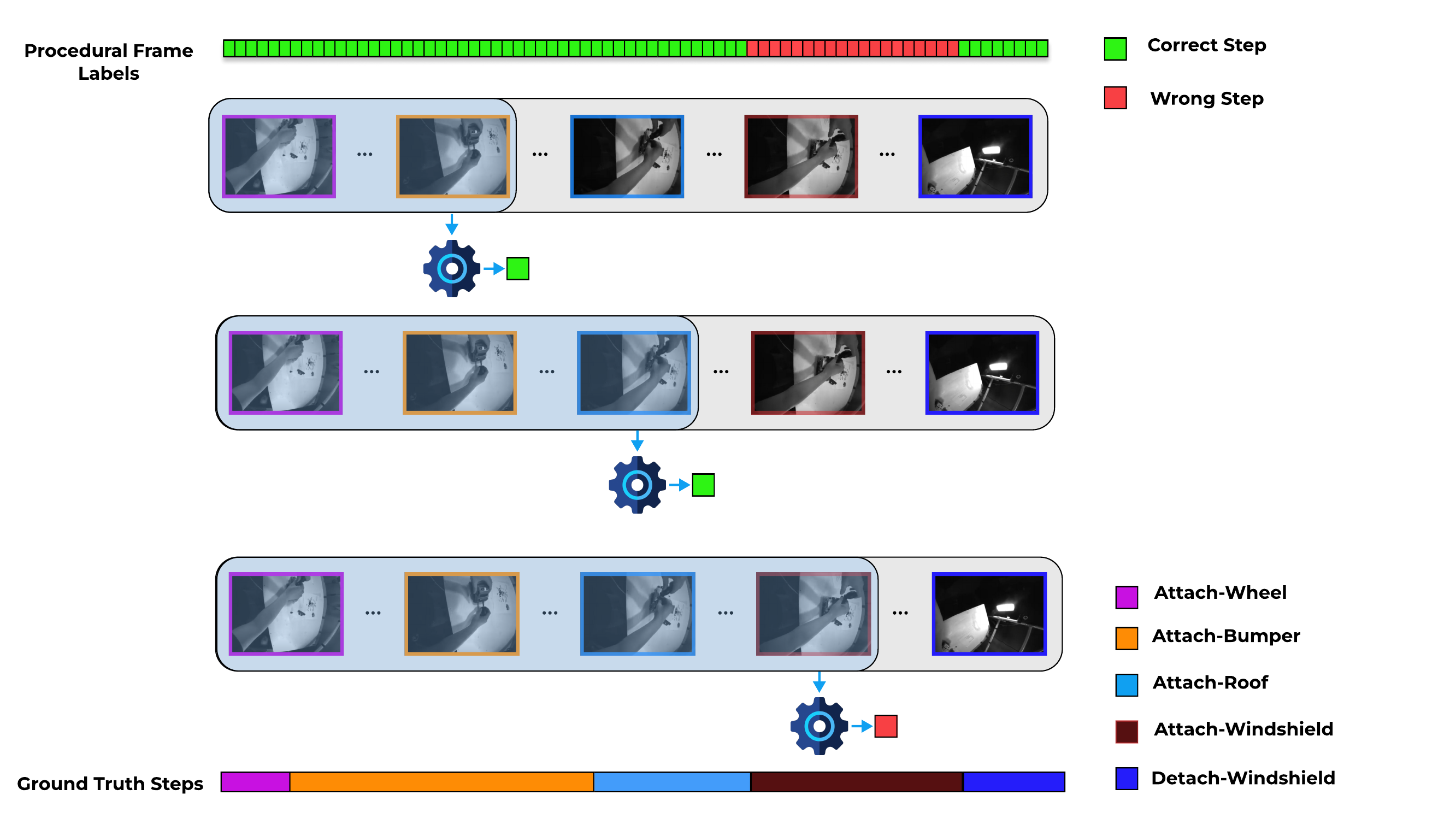} 
    \caption{
    Procedural Mistake Detection involves identifying errors within a procedural video. 
    Each procedure is composed of different steps that should be executed with a certain order.
    The aim is to develop a method capable of analyzing a video and determining whether each frame contains a mistake, e.g., a step that is not executed in the correct order. 
    Addressing this task is particularly useful when applied to wearable devices, as they allow for direct feedback to the person performing the task.
    In the Figure, TI-PREGO, depicted by a gear icon, takes as input the video sequence from time $0$ to $t-1$, classifying the current frame $t$ as either correct (\textcolor{teal}{green}) or a mistake (\textcolor{red}{red}).
    This process continues frame by frame until a  mistake is detected.
    } 
    
    \label{fig:Teaser}
\end{figure}

Advanced mistake detection models will become essential to ensure precision, safety, and efficiency in procedural applications. This demand has recently resulted in an increasing number of datasets~\cite{epicTent,sener2022assembly101,HoloAssist2023,Ghoddoosian_2023_ICCV,ding2023every,coin,ikea,IndustReal,cross_task,proceL_2019,howto100m_2019_miech,ragusa2023enigma51} and methodologies designed to advance procedure learning~\cite{POC_2022,Ghoddoosian_2023_ICCV,OadTR,zhong_2023,epicTent,coin,seminara2024differentiable} and error detection models~\cite{HoloAssist2023,ding2023every,IndustReal}. 
However, existing mistake detection approaches vary widely. Some methods emphasize action detection, identifying specific errors like missing steps, out-of-sequence actions~\cite{narasimhan2023learning} or incorrect ordering~\cite{seminara2024differentiable}, while other models bypass action detection altogether and monitor changes to the assembled object to ensure procedural accuracy~\cite{chen2020monitoring, candido2023image}. This variability in proposed approaches and their intended objectives highlights the need for more comprehensive evaluation frameworks and standardized benchmarks to assess and compare the effectiveness of different mistake detection strategies and for more olistic models, able to detect any deviation from the correct workflow.\\

An ideal Procedural Mistake Detection (PMD) model should demonstrate two critical capabilities: robustness to diverse mistake types and the ability to provide online, timely feedback. 
Regarding robustness, a reliable PMD system should be capable of identifying any mistake that may occur within a structured procedure, regardless of its type. To uphold this principle, in this paper we propose framing mistake detection as a One-Class Classification \added{(OCC)} problem, where the model is trained to learn the correct sequences of steps within a procedure and is then able to recognize any deviation from the correct workflow.
Within the OCC framework, during training, the PMD model is only exposed to ``normal'' procedures~\cite{Flaborea_2023_ICCV,zaheer2022generative} to learn the correct workflows, i.e. the sequences of performed steps that allow to correctly complete the task.
Conversely, during inference, the learned model should detect any deviation from the learnt workflow, deeming that step as an error. 
Further, it is worth noting that the OCC framework requires only binary labels (\textit{Correct} vs. \textit{Mistake}), which are easier to collect with respect to fine-grained labels,  
\added{such as} the error type\added{,} when labelling videos of procedural datasets.
Second, the model should provide online\footnote{We distinguish an online setup, which allows slight processing delays while enabling effective feedback, from a real-time setup, which requires immediate responses.} feedback quickly enough to allow users to correct mistakes promptly, minimizing the reinforcement of incorrect actions and reducing the risk of hazardous situations. 
\added{It is worth noting that the requirement for immediate feedback defines the online setting and imposes a strict causal constraint on the PMD models as they are allowed to leverage only past and present information for producing their decisions. In contrast, previous methods~\cite{Ghoddoosian_2023_ICCV,ding2023every,HoloAssist2023,lee2024error} adopt an offline framework, where the analysis is performed using complete pre-recorded procedures. This allows to employ bidirectional models that leverage both past and future context, achieving high accuracy in retrospective error identification for post-hoc analysis like performance grading, but making them unsuitable for live intervention. Thus, we argue that the processing approach must shift towards proactive error anticipation, where the model predicts the next correct step of the sequence to timely flag deviations.}
Besides these two pivotal abilities, the PMD system should leverage an egocentric perspective in procedural videos, enhancing the model's applicability by mirroring natural human perception, and allowing for seamless real-world deployment on wearable devices.\\

To conform to these requirements, we propose a dual-branch approach for detecting mistakes in procedural videos sequentially combining step recognition and anticipation and identifying mistakes as mismatches in predicted and observed actions. The first branch, the step recognition module, processes the input video online and classifies the current action being performed. The second branch, the step anticipation module, predicts the next step to perform based on the steps recognized by the step recognition module. We implement the anticipator branch as a Large Language Model (LLM), leveraging several prompting schemes to interrogate it and retrieve the next action to be performed. The model will then flag a step as mistaken when there is a divergence between recognized and anticipated actions, highlighting inconsistencies between the action executed by the operator and the expected steps.

Our contributions include (1) a novel benchmark for online mistake detection, encompassing two dataset (Assembly101-O and Epic Tent-O), (2) exploring and benchmarking multiple LLMs for the task of step anticipation aiming to select the most robust candidate for our anticipator module, (3) experimenting with various prompting techniques, including Automatic Chain of Thought (Auto-CoT), which enhances overall performance, reporting state-of-the-art results on Assemblty101-O and Epic Tent-O, (4) investigating different prediction aggregation techniques to improve mistake detection robustness and continuity. 
Extensive experimentation demonstrates the challenges and opportunities in dual-branch architectures and LLM-based step anticipation for open-set mistake detection. \\

This paper extends our previous conference work, PREGO~\cite{Flaborea_2024_CVPR}, by refining key components and methods to improve procedural mistake detection (see Fig.~\ref{fig:Teaser}).
Unlike our original study, which primarily reported classical Precision and Recall metrics, in this work, we adopt Balanced Precision~\cite{Wang2021OadTROA} to mitigate the extreme class imbalance inherent in egocentric mistake detection. Balanced Precision scales the False Positives by the ratio of positive‐to‐negative samples, ensuring that the penalty for mislabeling a correct step is placed on the same scale as the reward for correctly identifying a mistake. 
Specifically, we lead a comprehensive evaluation of several LLMs as step anticipators, increasing the state‑of‑the‑art results on Assembly101‑O and Epic Tent‑O~\cite{Flaborea_2024_CVPR}. We compare a broad range of LLMs such as Llama‑2~\cite{touvron2023llama}, Llama‑3~\cite{llama3}, Gemma~\cite{gemma}, and Mistral~\cite{mistral7B}, fine‑tune them using Low‑Rank Adaptation (LoRA), and experiment with prompting methods including Zero‑Shot, Few‑Shot, and Auto‑CoT~\cite{Zhang2022AutomaticCO}. 

Further, we revisit the recognizer branch’s frame-by-frame prediction mechanism, aiming to mitigate inconsistencies and reduce noise in its predictions. To deal with this, we extend beyond fixed-length window aggregation to explore alternative strategies that reduce noise and improve prediction accuracy.

\section{Related Works}\label{sec:relatedwork}
This section reviews the foundations of current approaches related to online detection of procedural mistakes,.
We first survey existing models and compare them across key design dimensions (Sec.~\ref{subsec:Procedural Mistake Detection}).
We then analyze the most relevant procedural video datasets (Sec.~\ref{subsec: Datasets of Procedural Egocentric Videos}), examine step recognition (Sec.~\ref{subsec: Step Recognition}) and anticipation (Sec.~\ref{subsec: Step Anticipation}) methods.
Finally, we discuss how recent advances in large language models enable symbolic reasoning (Sec.~\ref{subsec: Reasoning Tasks with Large Language Models}). Together, these components define the landscape for building effective online, open-set mistake detectors.

\subsection{Procedural Mistake Detection}
\label{subsec:Procedural Mistake Detection}
Procedural learning has seen notable progress due to the development of diverse datasets~\cite{coin,proceL_2019,cross_task,howto100m_2019_miech,ragusa2023enigma51}, which offer valuable insights into both structured~\cite{ikea,HoloAssist2023,Ragusa_meccano_2021} and unstructured tasks~\cite{epicTent,epickitchens_Damen_2018}. These datasets span a wide range of applications---from industrial assembly~\cite{Ragusa_meccano_2021,sener2022assembly101,IndustReal,ragusa2023enigma51} to everyday cooking activities~\cite{epickitchens_Damen_2018,breakfasts,salads}---yet the lack of a standardized benchmark for mistake detection has led to limited evaluation and a fragmented literature.

In Table~\ref{tab:mistake_methods}, we compare relevant models for procedural mistake detection across multiple dimensions: whether they operate in an open-set or one-class condition, whether they support online or offline detection, the data modalities they leverage (e.g., RGB frames, hand poses, eye gaze, keystep logic), the task they address, and the datasets used. Notably, models marked with an asterisk~(*) follow our protocol introduced in~\cite{Flaborea_2024_CVPR}, standardizing an online, open-set detection scenario for consistent evaluation.

Existing methods illustrate a variety of approaches. For instance, Ding et al.~\cite{ding2023every} employ knowledge graphs with textual transcripts, neglecting visual cues. Ghoddoosian et al.~\cite{Ghoddoosian_2023_ICCV} focus on action recognition, treating error detection as a semantic evaluation of segmentation results. Other works, such as \assembly~\cite{sener2022assembly101} and HoloAssist~\cite{HoloAssist2023}, apply error detection baselines offline, relying on pre-segmented videos and annotated labels. By contrast, \cite{IndustReal} emphasizes Procedure Step Recognition, focusing on correct step completion rather than partial activities, while EgoPER~\cite{lee2024error} integrates action segmentation with contrastive learning to detect previously unseen errors. Our prior work, PREGO~\cite{Flaborea_2024_CVPR}, uses RGB frames alongside symbolic reasoning to assess procedural correctness online, operating under an OCC framework and trained solely on error-free sequences~\cite{Flaborea_2023_ICCV,zaheer2022generative}. Finally, Seminara et al.~\cite{seminara2024differentiable} adopt a distinct approach, learning task graphs via Maximum Likelihood estimation of key-step sequences.

In contrast to prior approaches that rely on offline processing, pre-segmented videos, or textual-only cues, our method uniquely fuses real-time visual analysis with symbolic reasoning within an OCC framework. 
By learning solely from error-free sequences, our approach can detect known and unforeseen mistakes without being constrained to predefined error types. 
As indicated in Table~\ref{tab:mistake_methods}, we are the first to propose a fully online OCC protocol , allowing detection to be performed frame-by-frame without requiring prior action segmentation. 
This design enables a more robust, context-aware, and prompt detection of errors, addressing the key limitations observed in existing literature.

\subsection{Datasets of Procedural Egocentric Videos}
\label{subsec: Datasets of Procedural Egocentric Videos}
Previous investigations proposed datasets and benchmarks to support research in procedural mistake detection.
IndustReal~\cite{IndustReal} focuses on a single toy, resulting in the learning of only one procedure without annotations for procedural mistakes.
\assembly~\cite{sener2022assembly101} is a large-scale video dataset that provides frame-level mistake annotations. 
It features videos of actors assembling toys, with synchronized Ego-Exo views and annotated hand positions. 
%2 IndustReal
IndustReal~\cite{IndustReal} focuses on a single toy, resulting in the learning of only one procedure without annotations for procedural mistakes.
%3. epictent
\epic~\cite{epicTent} is a dataset that covers a different domain of unscripted actions that capture actors assembling a tent outdoors.
The participants exhibit varying levels of expertise and naturally make mistakes, which have been annotated.
%4 ATA
ATA~\cite{Ghoddoosian_2023_ICCV} is a procedural dataset created for offline mistake detection in assembling activities. 
However, it provides only video-level mistake annotations, limiting its usefulness for frame-based applications.
%5. HoloAssist
HoloAssist~\cite{HoloAssist2023} provides egocentric videos of individuals performing multiple manipulation tasks following expert instructions.
%6. Captain Cook 4D
CaptainCook4D~\cite{peddi2023captaincook4d} is tailored for evaluating procedural activities, specifically focusing on the culinary domain. 
%7. EgoPER
Notably, ~\cite{lee2024error} introduced a novel egocentric procedural error dataset consisting of videos depicting various errors within the cooking domain.

In this study, we propose a new benchmark building upon  the datasets from \cite{sener2022assembly101,epicTent} as they provide insights into procedural errors in two distinct settings: a controlled industrial environment and outdoor environments.

\begin{table*}[!t]
\centering
\caption{Comparison of relevant models in procedural mistake detection. In the modalities column, \textit{RGB} refers to RGB images, \textit{H} stands for hand poses, \textit{E} represents eye gaze and \textit{K} indicates keystep labels.
Our approach is the first to adopt an egocentric, one-class and online method for detecting procedural mistakes. (*) denotes that these models are based on our protocol defined in~\cite{Flaborea_2024_CVPR}.}

\label{tab:mistake_methods}
\resizebox{\textwidth}{!}{
\begin{tabular}{l|c|c|c|l|l|l}
\hline
\textbf{Model} & \textbf{Egocentric (Ego)} & \textbf{One-Class (OCC)} & \textbf{Online} & \textbf{Modalities} & \textbf{Task} & \textbf{Datasets} \\ \hline \hline
%1
Ding et al. \cite{ding2023every} - \textit{ArXiv '23} &  &  &  & \textit{K} & Mistake Detection & \assembly \cite{sener2022assembly101} \\
%2
Wang et al. \cite{HoloAssist2023} - \textit{ICCV '23} & \cmark &  &  & \textit{RGB}+\textit{H}+\textit{E} & Mistake Detection & HoloAssist \cite{HoloAssist2023} \\
%3
Ghoddoosian et al. \cite{Ghoddoosian_2023_ICCV} - \textit{ICCV '23} &  &  &  & \textit{RGB} & Unknown Sequence Detection & ATA \cite{Ghoddoosian_2023_ICCV}, CSV \cite{Qian_2022_CVPR} \\
%4
Schoonbeek et al. \cite{IndustReal} - \textit{WACV '24} & \cmark &  & \cmark & Multi & Procedure Step Recognition & IndustReal \cite{IndustReal} \\
%5
Lee et al. \cite{lee2024error} - \textit{CVPR '24} & \cmark & \cmark &  & RGB & Mistake Detection & EgoPER \cite{lee2024error}, HoloAssist \cite{HoloAssist2023}, ATA \cite{Ghoddoosian_2023_ICCV} \\
%6 trained
Seminara et al. \cite{seminara2024differentiable}
- \textit{NeurIPS'24} * & \cmark & \cmark & \cmark & \textit{RGB} & Mistake Detection & \textit{\assemblyO}, \textit{\epicO} \\
%7 not trained
PREGO - \textit{CVPR '24} * \cite{Flaborea_2024_CVPR} & \cmark & \cmark & \cmark & \textit{RGB} & Mistake Detection & \textit{\assemblyO}, \textit{\epicO} \\ \hline \hline
%8 not trained
\textbf{TI-PREGO} * & \cmark & \cmark & \cmark & \textit{RGB} & Mistake Detection & \textit{\assemblyO}, \textit{\epicO} \\ \hline

\end{tabular}}
\end{table*}

\subsection{Step Recognition}
\label{subsec: Step Recognition}

Step recognition is the process of identifying discrete actions within a structured procedural sequence.

Zhuang {\it et al.}~\cite{POC_2022} introduce an action segmentation model that employs an attention-based architecture coupled with a Pairwise Ordering Consistency (POC) loss function.
Moreover, they develop a weakly-supervised method that requires only the set of actions in a procedure as input, eliminating the need for labor-intensive frame-level annotations.
%2
In Shah {\it et al.} \cite{Shah_2023_STEPs}, a novel loss for self-supervised learning is combined with a clustering algorithm to detect key steps in procedural videos without using any labels.
%3 
\cite{zhong_2023} tackles the task by utilizing online instructional videos to learn actions and sequences without manual annotations, combining step recognition with a deep probabilistic model to account for step order and timing variability.
%4

An et al.~\cite{An2023MiniROADMR} introduced MiniROAD, the state-of-the-art online step recognizer, which leverages an RNN architecture and adjusts loss importance during training to perform online action recognition. 
The approach proposed in Shen et al.~\cite{shen2024progress} also highlights the importance of holistically understanding actions rather than treating each frame in isolation. In contrast to previous methods that rely on frame-by-frame detection—such as MiniROAD~\cite{An2023MiniROADMR}, which focuses on step recognition—our approach addresses the limitations of isolated frame-level analysis. While prior work has established a strong foundation in procedural step recognition, it often falls short in capturing the broader temporal context that is critical in real-world scenarios. To overcome this, we propose and evaluate several aggregation techniques that enhance the coherence of action recognition across sequences, thereby enabling more robust and context-aware mistake detection.

\subsection{Step Anticipation}
\label{subsec: Step Anticipation}
Step anticipation is the process of predicting the next action in a procedural sequence before it occurs. 

\cite{Abdelsalam_2023_gepsan} generates multiple possible natural language outcomes, leveraging pretraining on a text corpus to deal with the variability in future actions.
\cite{mascaro2023intention} employs a two-level hierarchical approach, integrating both high-level human intentions and low-level action sequences, improving action anticipation in the long term.
The pipeline consists of two models: the first one extracts intentions and classifies actions while the second one generates future actions, conditioning human intentions to narrow down the uncertainty set of future actions.
The framework presented in \cite{self_regulated} addresses future activity anticipation in egocentric videos by employing a contrastive loss to emphasize novel information and a dynamic reweighting mechanism that prioritizes informative past content. This approach improves video representation, leading to more accurate predictions of future activities.

Similar to AntGPT~\cite{Zhao2023AntGPTCL}, we leverage video frames to anticipate future actions using an LLM enhanced with Automatic Chain of Thought.
However, while previous approaches, such as AntGPT, require predefined segmentation of a long video into ordered, annotated short segments with corresponding action labels, our online setting processes each frame independently without prior segmentation. 

\subsection{Reasoning Tasks with Large Language Models}
\label{subsec: Reasoning Tasks with Large Language Models}
Large Language Models (LLMs), trained on vast datasets and equipped with numerous parameters, exhibit advanced capabilities beyond those of earlier language models~\cite{Wei2022EmergentAO}. 
They have demonstrated exceptional performance in both natural language processing tasks~\cite{touvron2023llama} and non-language tasks~\cite{brooks2023instructpix2pix, Wei2022EmergentAO, gupta2023visual}. 
The next-token prediction mechanism of LLMs closely parallels our action anticipation framework, as both aim to predict future outcomes based on accumulated data.

Previous research has shown that LLMs can generate semantically meaningful patterns~\cite{generalpatternmachines2023, Zhao2023AntGPTCL}, while other work~\cite{Wei2023LargerLM} has explored their in-context learning abilities with semantically unrelated labels, where there is no direct relationship between a token and its meaning. 
More recent work ~\cite{Pallagani2022PlansformerGS, gupta2023visual, Liang2022CodeAP, Feng2023LanguageMC, generalpatternmachines2023, Kim2024PALMPA, Ahn2022DoAI} has further investigated the ability of LLMs to function as \textit{In-Context Learners} (ICLs), meaning they can solve novel and unseen tasks relying only on a structured and informative prompt, without requiring additional fine-tuning. 
When provided with a query prompt that includes a context of input-output examples, LLMs can understand the problem and generate appropriate solutions within this framework.
LLMs as ICLs have been applied to a wide range of tasks, including planning~\cite{Pallagani2022PlansformerGS, Ahn2022DoAI}, programming~\cite{gupta2023visual, Liang2022CodeAP}, logical problem-solving~\cite{Feng2023LanguageMC} and symbolic reasoning~\cite{generalpatternmachines2023}.
In addition to ICL, other paradigms, such as Chain-of-Thought (CoT)~\cite{Wei2022ChainOT} and Automatic Chain-of-Thought (ACoT)~\cite{Zhang2022AutomaticCO}, have been employed to enable LLMs to reason through their responses step-by-step, improving their ability to plan and articulate their thought process explicitly.

Palm~\cite{Kim2024PALMPA} is the most similar to our anticipation module, as it employs a Socratic method~\cite{Zeng2022SocraticMC} that uses video features to predict subsequent actions with an LLM. 
Socratic Models are a framework that utilizes structured dialogue between pre-existing foundation models, each leveraging its unique capabilities based on its training data distribution. 
However, while PALM ~\cite{chowdhery2023palm} integrates an Action Recognition Model and a Vision-Language Model to generate text prompts for long-term action anticipation, we diverge by using a dual-branch architecture: one branch tasked to recognize the current action and the other leveraging an LLM for action anticipation focusing on next step prediction.
This schema offer a natural pipeline for error detection by comparing the predictions generated by the two branches

In our mistake detection pipeline, we integrate In-Context Learning and Automatic Chain of Thought, utilizing an LLM as part of our action anticipation branch. 
ACoT enables the model to automatically generate intermediate reasoning steps, breaking the problem-solving process into smaller, logical steps.
The approach helps the LLM to externalize its reasoning, enhancing its effectiveness in managing complex, sequential tasks.

% METHODOLOGY
\section{Methodology}\label{sec:background}
\begin{figure} 
    \centering
    \includegraphics [trim={3cm 0cm 3cm 0cm}, width=0.8\linewidth]{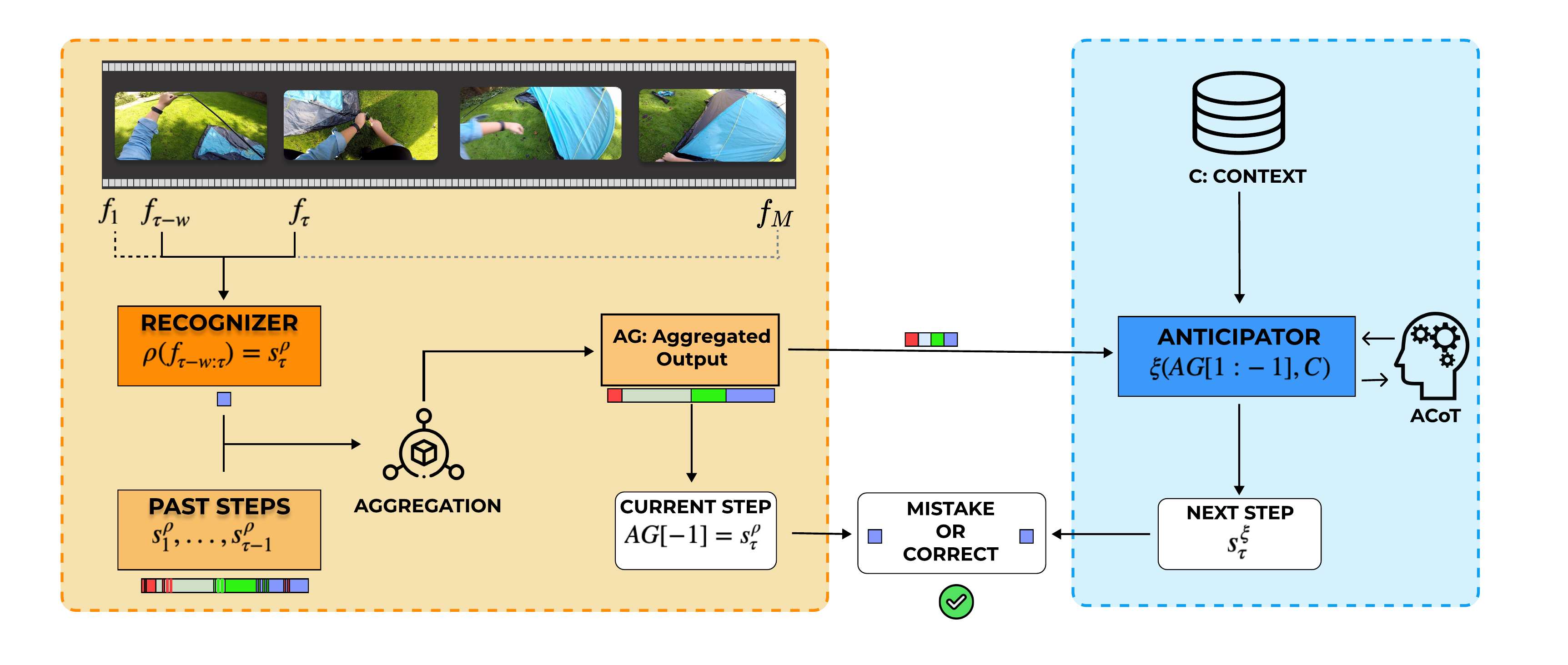}
    \caption{
     Our proposed model is based on two main components. The recognition module (orange) processes the input video in an online fashion and predicts actions observed at each timestep represented as a square. This prediction is combined with previous ones to minimize the noise generated by per-frame predictions. The last element of the aggregated output corresponds to the current step predicted by the recognition module, while the sequence without duplicates is provided to the anticipation module.
     The anticipation module (blue) reasons symbolically via a Large Language Model, utilizing automatic Chain of Thought (ACoT) reasoning to predict the future action based on past action history and a brief context such as instances of other step sequences. Mistakes are identified when the current action detected by the step recognition method differs from the one forecasted by the step anticipation module. 
     }
    \label{fig:conceptual}
\end{figure}

%TEASER
The proposed system leverages a dual-branch framework that integrates the recognition of procedural steps with anticipation modeling, as shown in Fig.~\ref{fig:conceptual}.
In the following sections, we elaborate on the problem formalization (Sec. \ref{sec3:prego}), present the branches for Step Recognition (Sec. \ref{sec3:miniroad}) and Step Anticipation (Sec. \ref{sec3:symb}) and finally illustrate the Mistake Detection procedure (Sec. \ref{sec3:mistake}).

\subsection{Preliminaries}
\label{sec3:prego}
Let us consider a finite collection of $N$ procedures $\{p_i\}_{i=1}^N$, where each procedure encodes a sequence of steps as $p_i = \{ s_k\}_{k=1}^{K_i}$ that represents the \textit{transcript} of $p_i$.
Here, $K_i$ varies depending on the specific procedure, and each step $s_k$ belongs to the set of all possible steps $\mathcal{S}$ in the dataset.
%$s_k \in \mathcal{S}= \{s | s \text{ is a possible step}\}$.
Furthermore, each procedure is represented by a set of videos $\{v_i\}_{i=1}^N$, which consist of frames $v_i=\{f_\tau\}_{\tau=1}^{M_i}$, with $M_i$ indicating the total number of frames in video $i$. 
Given a specific frame $f_{\tau}$ from video $v_{i}$, we:  
(1) classify the step $s_{\tau}$ corresponding to the frame $f_{\tau}$ (i.e., step recognition) and 
(2) predict the step $s_{\tau}$ that occurring at time $\tau$, relying solely on the previously recognized steps up to time $\tau-1$ (i.e., step anticipation).

Step recognition is handled by a module $\rho$, which takes the encoded frames of $v_{i}$ up to the $\tau$-th frame as input and outputs the recognized step $s^{\rho}_{\tau}$ (Sec.~\ref{sec3:miniroad}). 
Subsequently, all recognized steps $s^{\rho}_1, ..., s^{\rho}_{\tau-1}$, once aggregated, are fed into the module $\xi$, which handles the anticipation task by predicting the next step in the procedure, referred to as $s^{\xi}_{\tau}$(Sec.~\ref{sec3:symb}).

Finally, we compare $s^{\rho}_{\tau}$ with $s^{\xi}_{\tau}$ 
and deem the step as mistaken if there is a misalignment between the outputs from the two branches (Sec.~\ref{sec3:mistake}).
For clarity, the rest of this section will focus on a single procedure \(p\) associated with a video \(v\).

\subsection{Step Recognition}
\label{sec3:miniroad}
The step recognition module, denoted $\rho$, takes a window $W=\{{f_{{\tau}-w},..,f_{\tau}}\}$ with a fixed length $w$ of frames from $v$ to the current frame $\tau$ as input and is tasked with recognizing the last step performed $s^{\rho}_{\tau}$.
We leverage \added{MiniROAD}~\cite{An2023MiniROADMR} as $\rho$, which currently represents the state-of-the-art and is also computationally efficient in terms of GFlops and parameter count.

In this setup, the model predicts step $s_{\tau}$ by analyzing the frames ${f_{\tau-w },..,f_{\tau}}$. 

\begin{equation}
\rho( f_{\tau-w : \tau} )=s^{\rho}_{\tau}
\end{equation}

During training, the loss for $\rho$ is calculated using a cross-entropy loss, comparing the actual step $s_{\tau}$ with the predicted one $s^{\rho}_{\tau}$ as detailed in~\cite{An2023MiniROADMR}.

\begin{figure}[t]
    \centering
    \includegraphics[width= 1\linewidth]{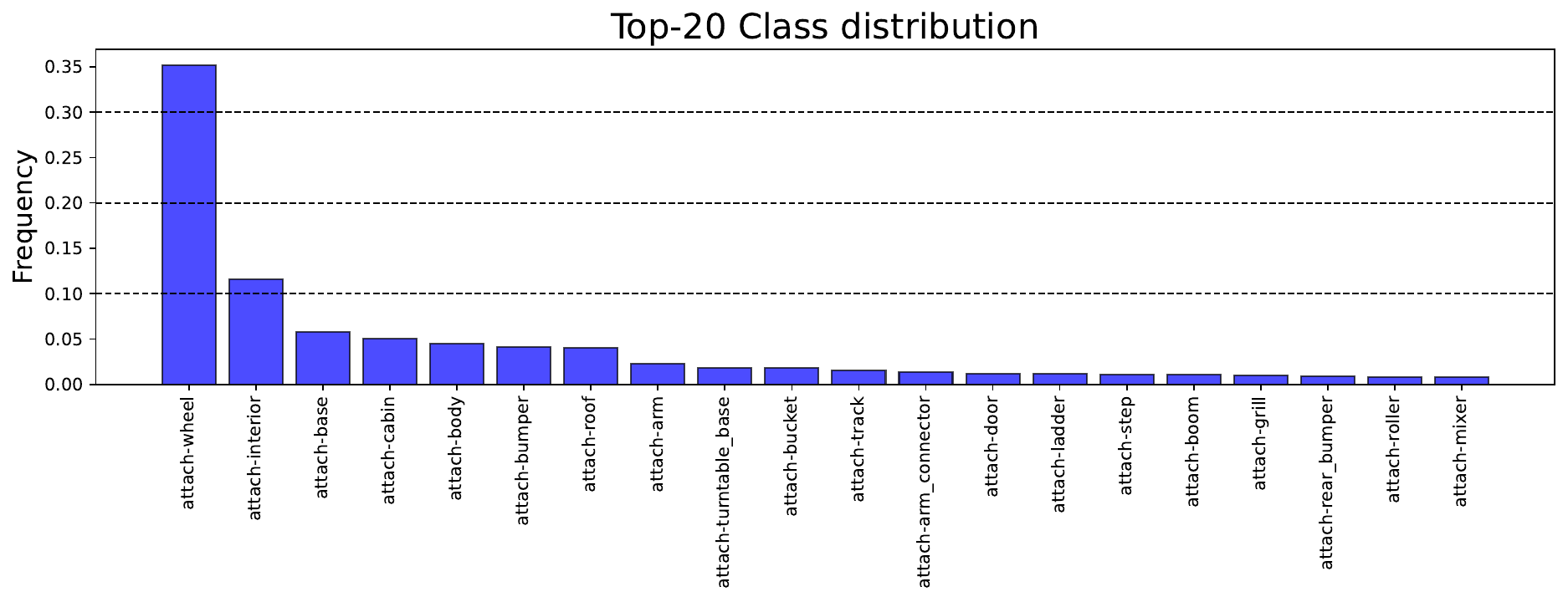}
    \caption{ \centering{Distribution of the 20 most frequent classes in  \assemblyO}}
    \label{fig:class cardinality}
\end{figure}

\paragraph{Step aggregation}
Despite being a state-of-the-art model, \added{MiniROAD}~\cite{An2023MiniROADMR} performs suboptimally on procedural datasets, primarily due to two key factors:
the shift in action distribution between the training
and testing phases, and the inherent class imbalance within the dataset.
The distribution shift occurs because the actions seen during training often differ from those encountered during testing, leading to poorer generalization.
Additionally, Assembly101 contains a significant imbalance in class frequencies  (see Figure~\ref{fig:class cardinality} for \assemblyO's split), where
common actions dominate the dataset while less frequent yet crucial procedural actions are underrepresented.

While MiniROAD~\cite{An2023MiniROADMR} produces per-frame predictions, the anticipator expects a sequence of distinct actions as input. A standard approach for transforming per-frame predictions into action sequences involves aggregating identical neighboring predictions into a single action segment.
However, MiniROAD’s performance issues introduce significant noise into the input sequence for the anticipator module, mainly due to inconsistencies in per-frame predictions, such as oscillating between similar actions, which leads to unstable action predictions and degrades results in subsequent stages.

To address this, we propose \added{three} frame aggregation strateg\added{ies} that efficiently converts per-frame predictions into coherent action sequences, aiming to reduce noise and
stabilize predictions. This results in a cleaner input for the anticipator module, dubbed $AG$ in Fig.~\ref{fig:conceptual}.

\begin{figure}
    \centering
    \includegraphics[width=\linewidth]{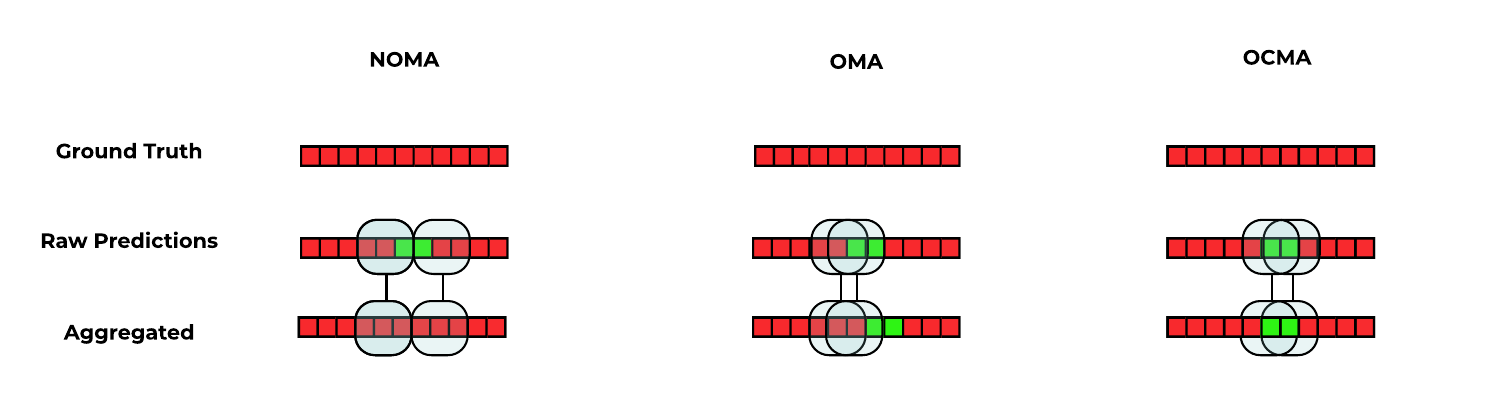}
    \caption{The three aggregation strategies operate differently: NOMA employs a non-overlapping window and replaces the entire window content with its mode. In contrast, both OMA and OCMA use overlapping windows with a sliding step of 1. However, OMA replaces the last frame of the window with its mode, whereas OCMA replaces the central frame. }
    \label{fig:ThreeAggregations}
\end{figure}

\added{
\paragraph{Non-Overlapping Mode Aggregation (NOMA)} The first aggregation strategy we propose (Figure \ref{fig:ThreeAggregations}) consists of a two-step process. In the first step, we calculate the mode for each non-overlapping sliding window of frames of a given length $L$.
\added{Since} the windows are non-overlapping, this approach introduces a slight delay in processing, which is a trade-off for improved prediction stability. The mode, representing the most frequent action label within that window, replaces all predictions in that window, ensuring consistency across the frames and removing the noise of smaller portions wrongly predicted.
In the second step, we apply the elimination of successive duplicates to remove consecutive repeated action labels, which prevents redundant predictions. 
This cleaned output serves as the input to the anticipation module, ultimately aiming to produce a more stable and accurate sequence of action predictions.
}

\added{
\paragraph{Overlapping Mode Aggregation (OMA) } 
The second aggregation strategy we propose (Fig. \ref{fig:ThreeAggregations})  employs a sliding window approach with a stride of $1$ frame, allowing for overlapping portions of frames.
In this method, we evaluate the mode within the overlapping window that includes the last frame and the previous $L-1$ frames. 
The prediction of the last frame is then substituted with this mode, which captures the most frequent action label in that localized area.
This method is designed with the intent to reduce delay, as it processes frames more dynamically compared to non-overlapping windows.
}

\added{
\paragraph{Overlapping Centered Mode Aggregation (OCMA)}
The third aggregation strategy (Fig. \ref{fig:ThreeAggregations}) uses a stride of 1 frame, similar to the previous approach.
To address the potential issue of the mode being incorrectly positioned within the window, we opt for substituting the prediction of the central frame rather than the final frame, with some corrections for the initial and final frames in the video. 
This adjustment ensures that the most frequent action label is applied to a frame that is more representative of the overall context, thus enhancing the accuracy of the prediction.
} \\

NOMA
uses nonoverlapping windows, while the other two employ overlapping windows. These two overlapping methods differ based on which frame is selected for imputing the aggregation result: the last frame (Overlapping Mode Aggregation, OMA) or the central frame (Overlapping Central Mode Aggregation, OCMA) (See Figure~\ref{fig:ThreeAggregations}).
All approaches are designed to reduce noise and eliminate redundant consecutive predictions, ensuring that the resulting sequence aligns closely with the actual procedure measured by the Levenshtein Similarity. 

In Sec.~\ref{sub:aggr}, we \added{assess,} compare and discuss in details the \added{impact of each} aggregation strateg\added{y}
and find NOMA to be the most effective, which we therefore select \added{for our proposal}.

\subsection{Step Anticipation}
\label{sec3:symb}
We employ an LLM as our $\xi$ module for next-step prediction, using as input transcripts from procedural video.
Our step anticipation operates without domain specific training or fine-tuning, leveraging only the reasoning from \textit{Automatic Chain of Thought (ACoT)}~\cite{Zhang2022AutomaticCO} (see Fig.~\ref{fig:conceptual}) and few-shot prompting method.
These prompts consist of three components: the system prompt, the ACoT prompt, which enables reasoning, and the contextual transcripts $C$ together with the current aggregated \textit{sequence} $AG$. 
$C$ is derived from similar transcripts (i.e., different videos for the object we are considering) to provide the LLM with information about typical step sequences and their order (see following examples).
\textit{Sequence} $S^{\rho}_{\tau}$, includes the current sequence of actions up to the current frame $f_{\tau-1}$, as they have been detected by our module $\rho$, i.e., $S^{\rho}_{\tau} = [s^{\rho}_{1},...,s^{\rho}_{\tau-1}]$, where $s^{\rho}_{i}$ represents the action detected by the module $\rho$ at frame $i$. $S^{\rho}_{\tau}$ then undergoes our aggregation strategy, which results in $AG$.  

The prompting occurs in two stages. First, we employ an ACoT mechanism of $\xi$ as a first step. This ACoT process uses context $C$, aggregation sequence $AG$, and an additional prompt to explicitly stimulate reasoning (See \textit{AcoT Prompt} in the following scheme). 
Then, we use the output of the intermediate step along with $C$ and $AG$ to predict the next most probable action.
By breaking down the task into smaller logical steps, the LLM can leverage the contextual transcript $C$ and the current action sequence $AG$ to generate intermediate reasoning steps, bridging the gap between observed and future actions $s^{\xi}_{\tau}$.

\vspace{0.5cm}
\begin{samepage}
\begin{systemframe}
    \textit{System Prompt: } \\
    Below is an instruction that describes the task, paired with an input. Write a response that appropriately completes the request.
\end{systemframe}
\end{samepage}

\vspace{0.5cm}
\begin{samepage}
\begin{acotframe}
\textit{ACoT Prompt:} \\
Let's analyze each action in detail:
\begin{enumerate}
    \item For the action ``attach the cabin", consider the immediate effect and long-term consequences. 
    \item Repeat this analysis for each action in the sequence.
    \item Identify patterns and causal relationships between the actions.
\end{enumerate}
Now, let's proceed with the analysis step-by-step:
\end{acotframe}
\end{samepage}

\vspace{0.5cm}
\begin{samepage}
\begin{contextframe}
    \textit{Contextual Prompt:} \\
    \textbf{Toy:} ``bulldozer"\\
    \textbf{Input Sequence:}\\
    attach-cabin, attach-body, attach-track, attach-blade\\
    \textbf{Next Symbol:}\\
    attach-figurine
\end{contextframe}
\end{samepage}

\vspace{0.5cm}
\begin{samepage}
\begin{sequenceframe}
    \textit{Current sequence:} \\
    attach-door
\end{sequenceframe}
\end{samepage}

\subsection{Mistake Detection}\label{sec3:mistake}
Finally, we analyze the output of the two modules to identify procedural errors. 
Specifically, we classify steps as correct when the outputs of both modules match, while we label cases as errors when the outputs differ. 
That is:
\begin{equation}
\begin{cases}
    s^{\rho}_{\tau} \neq s^{\xi}_{\tau} & \text{MISTAKE} \\
    s^{\rho}_{\tau} = s^{\xi}_{\tau} & \text{CORRECT} \\
\end{cases}
\end{equation}

% EXPERIMENTS
\section{Experiments}

This section describes the benchmark datasets, adopted evaluation metrics and the main experimental results.
We introduce {\it \assemblyO} and {\it \epicO} as adaptations of the original datasets~\cite{sener2022assembly101,epicTent}, detailing the selected labeling criteria for online benchmarking, a novel configuration of training and test splits to accommodate open-set procedural mistakes.
We define our proposed online metrics in Sec.~\ref{res:metrics}.
Finally, we introduce the main experimental results in Sec.~\ref{sec:experiments}, whereas further insights that motivate our design choices are provided through the ablation studies in Sec.~\ref{sec:Ablation}.

\subsection{Datasets}
To align with real-world applications, this work focuses on Egocentric videos.
 
\subsubsection{\assemblyO}
\label{sec4.1:Assembly101}
\assembly~\cite{sener2022assembly101} is a large-scale video dataset containing 362 videos of assembly and disassembly tasks involving 101 types of toy vehicles, recorded from eight multiple static and four egocentric cameras. 
It includes extensive annotations at various levels of granularity, with over 100K coarse action segments and 1M fine-grained action segments and 18M 3D hand poses. The dataset supports diverse challenges, including action anticipation and segmentation, mistake detection and 3D pose-based action recognition.

\begin{figure}
    \centering
    \includegraphics[width=1\linewidth]{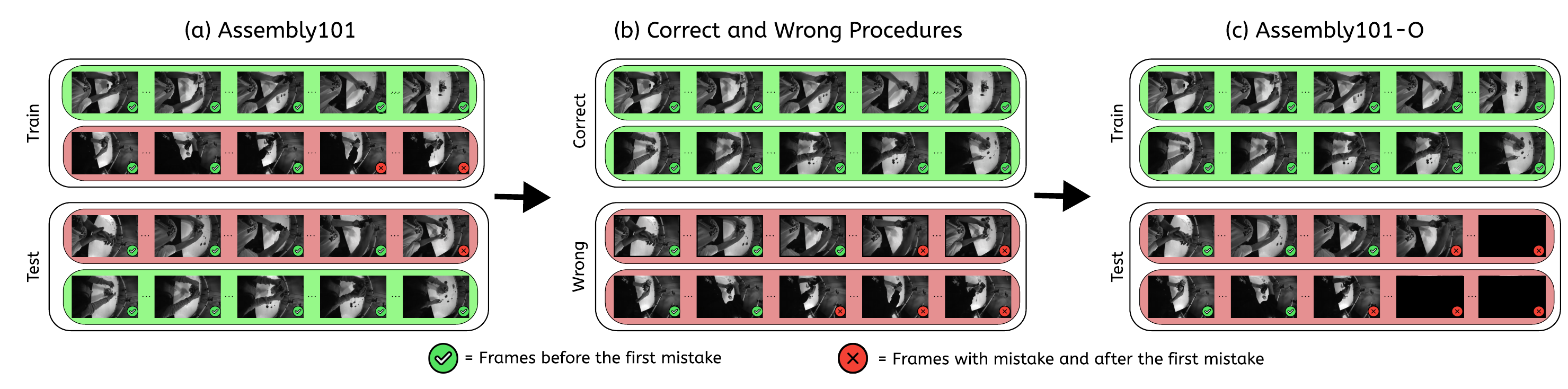}
    \caption{The modification occurs in two steps:
    starting from Assembly101 (a), first, a new train/test split assigns all correct procedures to the training set, reserving videos with mistakes for testing (b). Second, the lengths of procedures containing mistakes are adjusted by trimming them to the first mistake, preventing the creation of corrupted sequences (c). This setup enables models to learn correct sequences within a one-class classification (OCC) framework, treating any deviation as a mistake and ensuring a balanced test set for effective mistake detection.
    }
    \label{fig:dataset}
\end{figure}

\myparagraph{\assembly for Online and Open-Set Mistake Detection (\textit{Proposed})}
We introduce a novel split of the dataset~\cite{sener2022assembly101} tailored explicitly for online, open-set mistake detection. 
\assemblyO introduces two main modifications to~\cite{sener2022assembly101} as we show in Figure \ref{fig:dataset}: a new train/test split and revised procedure lengths. 

The new split assigns all correct procedures to the training set while reserving videos with mistakes for testing. 
This adjustment enables models to learn sequences characteristic of correct procedures in a one-class classification (OCC) framework, avoiding bias from specific mistake types during training.

The second adjustment limits each procedure’s evaluation to the point where a mistake caused by an incorrect step that distrupts the workflow:
evaluating beyond the point of procedural compromise creates discrepancies between the training and test set procedures, hindering effective step recognition and anticipation.

\subsubsection{\epicO}
\label{sec4.1:EPIC-tent}

\epic is a dataset of egocentric videos capturing the outdoor assembly of a camping tent. 
It was collected by 24 participants who wore two head-mounted cameras (a GoPro and a SMI eye tracker) while performing the task, resulting in 5.4 hours of recorded video. 
The dataset includes annotations for action labels, task errors, participant self-rated uncertainty and gaze positions, capturing the variability and complexity of the task as participants interacted with non-rigid objects such as the tent, guylines, instructions and tent bag, and displayed varying proficiency and confidence levels.

\myparagraph{\epic for Online and Open-Set Mistake Detection (\textit{Proposed})}
We propose a novel split for the \epic dataset~\cite{epicTent} tailored for open-set mistake detection. The dataset includes annotations for nine distinct mistake types, though only four of these categories —``{\it order}", ``{\it omit}", ``{\it correction}" and ``{\it repeat}" — are considered procedural mistakes, as they represent deviations that compromise the task's procedural integrity. 
Other categories such as ``{\it slow}", ``{\it search}", ``{\it misuse}", ``{\it motor}" and ``{\it failure}" are not procedural errors, as they do not disrupt the task’s progression.

Unlike \assembly~\cite{sener2022assembly101}, \epic is designed explicitly for supervised error detection with each recorded procedure containing at least one mistake, making it incompatible with the split procedure proposed for \assemblyO. 
However, \epic provides %frame-level 
confidence scores from each participant, reflecting their self-assessed uncertainty during task execution. 
Therefore, we define a custom split strategy where videos from higher-confidence participants form the training set.
In contrast, participants showing more significant uncertainty, and thus potentially prone to more errors, make up the test set. 
In practice, the videos that form the test set have a median confidence score under 0.6, while the others form the train set. 
Only 22 videos have the confidence score annotations, while the remaining 7 do not and are assigned to the test set.
This partition results in 14 training videos and 15 test videos.

This strategy is particularly promising for real-world applications where accurately labeling erroneous frames may be challenging or training a mistake detector can start soon after recording without requiring full annotation completion. 

Videos in the test set are trimmed to the frame containing the first procedural mistake as in \assemblyO, while videos representing correct procedures remain unaltered.

\subsection{Evaluation Definition}
\label{res:metrics}

To assess the performance of our procedural mistake detection model, we consider True Positives (TP) as the count of errors correctly identified, and True Negatives (TN) as the count of correctly performed steps that are not flagged as errors. Since, in ego-centric mistake detection, the positive class (actual mistakes) is typically much smaller than the negative class (correct steps), a straightforward application of classical Precision will tend to underestimate the impact of False Positives (FP). In particular, even a small FP rate can yield a large absolute number of FP simply because there are so many more negative (non-error) samples than positive (error) samples.

To mitigate this imbalance, we utilize \textit{Balanced Precision} \cite{Wang2021OadTROA}, denoted \(\mathrm{Prec}_{b}\), by scaling the contribution of each False Positive by the ratio of positive-to-negative samples. Concretely, if \(N_{+}\) is the total number of positive (error) samples and \(N_{-}\) is the total number of negative (correct) samples, we define:
\begin{equation}
\mathrm{Prec}_{b} \;=\; 
\frac{\mathrm{TP}}
     {\mathrm{TP}\;+\;\Bigl(\tfrac{N_{+}}{N_{-}}\Bigr)\,\mathrm{FP}}\,.
\end{equation}
Thus, FP is down-weighted by the factor \(\tfrac{N_{+}}{N_{-}}\), ensuring that the penalty for mistakenly labeling a correct step as an error is placed on the same scale as the reward for correctly identifying a mistake. When \(N_{+} \ll N_{-}\), this adjustment prevents a handful of FPs from dominating the denominator and driving classical Precision toward zero.

We retain \textit{Recall} (also known as Sensitivity for the positive class) in its standard form, since Recall inherently measures the proportion of actual mistakes that the model retrieves:
\begin{equation}
\mathrm{Recall} \;=\; \frac{\mathrm{TP}}{\mathrm{TP} + \mathrm{FN}}\,.
\end{equation}
Because Recall does not depend on the cardinality of the negative class, it is not directly skewed by the negative-class majority and therefore needs no further adjustment.

Finally, we define a \textit{Balanced F1 Score}, denoted \(\text{F1 score}_{b}\), which is the harmonic mean of \(\mathrm{Prec}_{b}\) and \(\mathrm{Recall}\):
\begin{equation}
\text{F1 score}_{b} \;=\; 
2 \,\frac{\mathrm{Prec}_{b} \times \mathrm{Recall}}
         {\mathrm{Prec}_{b} + \mathrm{Recall}}\,.
\end{equation}

These balanced metrics specifically assess the model’s ability to distinguish correct from incorrect steps, without being influenced by heavily skewed settings. 
By adjusting the impact of false positives based on the imbalance between positive and negative classes, \(\mathrm{Prec}_{b}\) and \( \text{F1 score}_{b}\) provide a fairer assessment of mistake detection when the negative class vastly outnumbers the positive class. 

\subsection{Experiments}
\label{sec:experiments}

\subsubsection{Compared Methods}
To estimate the effectiveness of our model, we evaluate its performance by comparing it against several baseline models based on the metrics presented in Sec.~\ref{res:metrics}:

\myparagraph{One-step memory}
The One-step memory baseline stems from a {\it transition matrix} that considers only the correct procedures. 
Specifically, given the set of actions $\mathcal{A}$ in the training set with $|\mathcal{A}|=C$, we define a transition matrix $M\in\mathbb{R}^{C\times C}$, which records, in position $(l,m)$, the number of occurrences where action $m$ follows action $l$. 
During evaluation, an action in the test split that does not correspond to a transition recorded in this training set matrix is labeled as a {\it mistake}. This baseline provides a simple method for recognizing deviations from standard procedural transitions.

\myparagraph{OadTR for mistake detection}
The work in \cite{OadTR} presents a framework for online action detection called OadTR, which uses a Vision Transformer to capture video clips' temporal structure and context. The framework consists of an encoder-decoder architecture: the encoder processes historical observations to output a task token that represents the current action, while the decoder refines this task token by incorporating information from anticipated future clips. In the context of procedural error detection, we consider a mistake to have occurred if the output from the encoder does not match the output from the decoder, indicating a divergence between observed and expected actions.

\myparagraph{BERT-based Mistake Detection}
We also leverage the capabilities of BERT~\cite{Devlin2019BERTPO}, using its specific [CLS] token to predict whether an action sequence is correct or erroneous.
More specifically, we fine-tune BERT with a next-token-prediction task, training the model to determine whether step B logically follows step A within a given procedure. In our setting, each step is represented as a set of two words, such as \textit{attach wheel}, which describes coarse-level actions. During evaluation, BERT is presented with pairs of actions and predicts whether the sequence of those actions is correct. 
The rationale of BERT lies in its pre-training on a vast corpus of text, which helps it understand contextual relationships between procedural steps.

\myparagraph{Learned Task Graph Representations}
We also evaluate an approach based on explicit task graph representations \cite{seminara2024differentiable}. Procedural activities consist of sequences of key steps that work toward specific goals, making task graphs a valuable, human-understandable representation for capturing the partial ordering of these steps. 
Unlike traditional methods that rely on hand-crafted procedures to extract task graphs from videos, this approach uses maximum likelihood optimization of edge weights to directly learn task graphs, allowing gradient-based learning that can integrate seamlessly with neural networks. 
While effective, this baseline is not directly comparable to our method, as it requires access to the entire dataset during training to extract the task graph, whereas our approach relies only on few-shot inference for anticipating actions. We refer to the ``Direct Optimization'' method trained with the ``Task-Graph Maximul Likelihood'' loss proposed in~\cite{seminara2024differentiable} as TGML-DO.

\subsubsection{ Results}
\label{res:Sequence-Level Results}

Table~\ref{tab:principale} presents the main results in terms of \(\mathrm{Precision}_{b}\), \(\mathrm{Recall}\), and \(\mathrm{\text{F1 score}}_{b}\).

In the upper section of the table, we report results obtained using an oracle step-recognizer, which serves as an upper bound. 
Here, the recognition module perfectly aligns with the ground truth, ensuring that errors stem solely from the anticipation module. 
The One-step memory method performs poorly as it relies only on the preceding action.
Even if BERT adopts a more abstract reasoning strategy, it suffers from a conservative bias that leads to very low recall, ultimately resulting in a significantly lower \(\mathrm{\text{F1 score}}_{b}\).
PREGO, on the other hand, incorporates symbolic reasoning to capture richer contextual information, with TI-PREGO$_{Llama}$ achieving the highest performance across most metrics.
\added{To ensure a fair comparison, we also evaluated the baseline PREGO model with Llama 3.1. As shown in Table~\ref{tab:principale}, TI-PREGO (74.7\% $\mathrm{\text{F1 score}}_{b}$) outperforms the PREGO baseline (67.5\% $\mathrm{\text{F1 score}}_{b}$) on \assemblyO, confirming that our methodological improvements are the primary driver of the performance gain.}

The TGML-DO method \cite{seminara2024differentiable}, which learns task graph representations through gradient-based optimization, also demonstrates strong performance.
However, unlike our anticipator, which operates in a few-shot inference setting, TGML-DO checks whether the observed action is correct by referencing domain-specific procedural knowledge mined from the entire dataset during training. 
This distinction likely contributes to the superior generalization of our model, particularly when using predicted steps instead of oracle labels in the \assemblyO dataset.
Our method proves to be more robust to procedural variations and better suited for handling previously unseen cases.

In the lower section of the table, we evaluate models using state-of-the-art step recognizers instead of an oracle. 
The OadTR model, designed for offline mistake detection, shows limited performance due to its reliance on fixed video segmentation, resulting in a low \(\mathrm{\text{F1 score}}_{b}\) of 26.7\% on Assembly101-O and 28.3\% on Epic-tent-O.
By contrast, combining MiniRoad with large language models for step anticipation leads to consistent gains.
When paired with GPT-3.5, PREGO achieves a 34,7\% absolute improvement in \(\mathrm{\text{F1 score}}_{b}\) over OadTR on Assembly101-O. 
The use of DeepSeek-R1 yields similar gains.
The best performance is obtained by TI-PREGO combined with Llama 3.1, which achieves an \(\mathrm{\text{F1 score}}_{b}\) of 74.7\% on Assembly101-O and 57.8\% on Epic-tent-O, outperforming all other configurations. 
\added{This direct comparison, using the same Llama 3.1 model, confirms our method's advantage: TI-PREGO (67.4\% $\mathrm{\text{F1 score}}_{b}$) outperforms the PREGO baseline (67.1\% $\mathrm{\text{F1 score}}_{b}$) on \assemblyO, with a similar trend on \epicO (34.6\% vs. 33.7\%).}
This highlights the robustness of Llama 3.1 in modeling procedural structure and anticipating upcoming steps, particularly in open-set, real-world mistake detection scenarios.
 
A more detailed discussion on the effectiveness of LLMs for step anticipation is provided in Section~\ref{exp:anticipator}.

\begin{table*}[!t]
\caption{
    A comparative assessment between Ours and the chosen baseline methods is conducted to detect procedural mistakes using the \assemblyO~and \epicO~datasets.
    } 
\label{tab:principale} 
 
\resizebox{\linewidth}{!}{
 \begin{tabular}{l|c|c|ccc|ccc} 
 \toprule
  & \multicolumn{2}{c|}{\textbf{}} & \multicolumn{3}{c|}{\textbf{\assemblyO}}  & \multicolumn{3}{c}{\textbf{\epicO}} \\ 
  & \multicolumn{1}{c|}{\textbf{Step Recog.}} & \multicolumn{1}{c|}{\textbf{Step Antic.}} &  \multicolumn{1}{c}{\textbf{Precision\textsubscript{b}}} & \multicolumn{1}{c}{\textbf{Recall}}  & \multicolumn{1}{c|}{\textbf{F1 score\textsubscript{b}}} & \multicolumn{1}{c}{\textbf{Precision\textsubscript{b}}} & \multicolumn{1}{c}{\textbf{Recall}} & \multicolumn{1}{c}{\textbf{F1 score\textsubscript{b}}}\\ 
 \hline

 \multirow{1}{*}{One-step memory} &  \textit{Oracle} &  & 43.87 & 30.7 & 36.2 & 42.1 & 26.6 & 32.6\\
 
 \multirow{1}{*}{BERT \cite{Devlin2019BERTPO}} &  \textit{Oracle} &  & 93.2  & 20.0  & 32.9 & 71.6 & 5.6 & 10.4 \\

 \multirow {1}{*}{{PREGO}~\cite{Flaborea_2024_CVPR}} &  \textit{Oracle} &  \textit{GPT-3.5} & 51.4 & 87.5 & 64.7 & 25.6 & 90.6 & 39.9 \\ 
 
 \multirow{1}{*}{PREGO~\cite{Flaborea_2024_CVPR}} &  \textit{Oracle} & \textit{DeepSeek-R1} & 51.7 & 98.3 & 67.8 & 23.8 & \textbf{100} & 38.4 \\
 \multirow{1}{*}{\added{PREGO}~\cite{Flaborea_2024_CVPR}} & \textit{Oracle} & \textit{Llama 3.1} & \added{56.8} & \added{89.6} & \added{69.5} & \added{28.0} & \added{92.0} & \added{42.9} \\
 \multirow{1}{*}{\textit{TI-PREGO}} &  \textit{Oracle} & \textit{Llama 3.1} & \textbf{60.4} & \textbf{97.8} & \textbf{74.7} & \textbf{40.7} & \textbf{100} & \textbf{57.8} \\
 \hline
 \multirow{1}{*}{\textit{TGML-DO~\cite{seminara2024differentiable}}} & \textit{Oracle}&  & 77.82 & 90.4 & 83.64 & 91.24 & 27.3 & 42.03 \\
 \multirow{1}{*}{\textit{TGML-DO~\cite{seminara2024differentiable}}} & \textit{MiniRoad~\cite{An2023MiniROADMR}}& & 53.77 & 37.3 & 44.05 & 96.55 & 14.1 & 24.60\\
 \hline
 
 \multirow{1}{*}{OadTR for MD~\cite{OadTR}}  &  \textit{OadTR~\cite{OadTR}} &  \textit{OadTR~\cite{OadTR}} & 50.9  & 18.1 & 26.7    & 40.4 & 21.7 & 28.3\\
 
 \multirow{1}{*}{{PREGO}~\cite{Flaborea_2024_CVPR}} & \textit{MiniRoad~\cite{An2023MiniROADMR}} &   \textit{GPT-3.5}  & 50.2 & 75.8 & 61.4 & 19.5 & 73.3 & 29.4 \\
 
 \multirow{1}{*}{PREGO~\cite{Flaborea_2024_CVPR}} &  \textit{MiniRoad~\cite{An2023MiniROADMR}} & \textit{DeepSeek-R1} & 51.2 & 96.5 & 65.7 & 20.3 & \textbf{93.3} & 33.6 \\
 \multirow{1}{*}{\added{PREGO}~\cite{Flaborea_2024_CVPR}} & \textit{MiniRoad~\cite{An2023MiniROADMR}} & \textit{Llama 3.1} & \added{51.3} & \added{96.8} & \added{67.1} & \added{20.6} & \added{93.3} & \added{33.7} \\
 
 \multirow{1}{*}{\textit{TI-PREGO }} & \textit{MiniRoad~\cite{An2023MiniROADMR}} &  \textit{Llama 3.1} & \textbf{51.5}  & \textbf{97.2}  & \textbf{67.4} & \textbf{21.2} & \textbf{93.3} & \textbf{34.6} \\
 \hline
 \end{tabular}
 }

\end{table*}

% ABLATION
\section{Ablation}
\label{sec:Ablation}

In the following subsections, we investigate the application of large-language models as step anticipators~\ref{exp:anticipator}. This subsection covers our model selection process, a detailed analysis of different prompting methods like Zero-Shot, Few-Shot, and Automatic-Chain-of-Thought (ACoT), and the effects of fine-tuning and speed analysis. 
Together, these experiments illustrate the strengths and challenges of using LLMs for action anticipation and highlight the trade-offs between accuracy, reasoning complexity and computational efficiency.

We also present a comprehensive evaluation of the performance of our system and the various strategies used to improve action prediction in assembly tasks. 
We introduce our Aggregation Strategy~\ref{sub:aggr}, detailing three methods to convert noisy per-frame predictions into coherent action sequences. We discuss the design of each strategy and their trade-offs, such as delay versus stability, and assess their effectiveness using the Levenshtein similarity metric.

\subsection{Large Language Models as Step Anticipators}
\label{exp:anticipator}
This section comprehensively analyzes how an action anticipation based on large language models (LLMs) operates using various techniques and methodologies. We begin by evaluating the performance of several LLMs using in-context learning with few-shot examples (Sec. \ref{exp:llm_selection}), which allows us to determine which model performs best in this setting. Once the most effective LLM is identified, we investigate different prompting strategies, starting from a system prompt's impact, including zero-shot, few-shot, and Automatic-Chain-of-Though prompting (Sec \ref{exp:prompt_analysis}).
Furthemore, we explore the potential improvements by fine-tuning each prompting method to enhance the model's performance (Sec. \ref{exp:finetuning}). Lastly, we perform a speed analysis, comparing different prompting methods (Sec. \ref{exp:speed_analysis}). 

All experiments use ground truth labels of the \assemblyO dataset as the output of the predictor's branches. 
This approach ensures that our performance analysis remains objective and untainted by errors from the step recognition branch, allowing for a more accurate evaluation of the step anticipator's capabilities.

\subsubsection{Model Selection}
\label{exp:llm_selection}
In this analysis, we evaluate the performance of various large language models (LLMs), as shown in Table~\ref{table:llm_selection}, using a few-shot context. The results reveal a clear performance hierarchy among the tested models. In particular, \textbf{LLAMA 3.1 8B}, an open-source model less computationally expensive during inference, demonstrates robust performance (see Table~\ref{tab:principale}). LLAMA 3.1 8B achieved the highest \(\mathrm{\text{Precision}}_{b}\) at 56.70\% and the best \(\mathrm{\text{F1 score}}_{b}\)of 69.4\%, significantly outperforming its counterparts.

The superior performance of LLAMA 3.1 8B could be attributed to several factors, including potential advancements in its training methodology, the quality and diversity of its training data, and possible optimizations for general task comprehension.  
A pattern observed across all models is the substantial disparity between \(\mathrm{\text{Precision}}_{b}\) and Recall scores. 
The consistently high Recall scores indicate that these models identify relevant assembly steps effectively, but the corresponding drop in \(\mathrm{\text{Precision}}_{b}\) suggests they also introduce a non-negligible number of spurious steps, underscoring the need for strategies, such as refined prompt engineering.

\begin{table}[!ht]
\caption{Results of different LLMs on \assemblyO.}\label{table:llm_selection}
\centering
\begin{tabular*}{\textwidth}{@{\extracolsep\fill}lccc}
\toprule
Model & Precision\textsubscript{b} & Recall & F1\textsubscript{b} \\
\midrule
Mistral 7B   & 53.4   & 88.9  & 68.1  \\
Gemma 9B     & 51.5    & 85.6  & 65.5  \\
Phi 3 medium 4k instruct & 54.0    & 88.7  & 68.4  \\
LLAMA 2 7B   & 54.0  & 94.0 & 68.6  \\
LLAMA 3 7B   & 53.8    & 87.3  & 67.3  \\
DeepSeek-R1  & 51.5    & \textbf{97.8}  & 67.5  \\
LLAMA 3.1 8B & \textbf{56.7}  & 91.3  & \textbf{69.4} \\
\botrule
\end{tabular*}
\end{table}

\subsubsection{Prompt Analysis}
\label{exp:prompt_analysis}

\paragraph{System prompt.}
The results in Table~\ref{tab:system_prompt} highlight the positive impact of combining Automatic-Chain-of-Though (ACoT) reasoning with Few-Shot (FS) examples on the performance of LLAMA 3.1 8B across different modalities in the \assemblyO dataset. In all prompting scenarios, the system prompt is written as: 

\vspace{0.5cm}
\begin{samepage}
\begin{systemframe}
    \textit{System: } I am going to provide an input sequence that represents a sequence of actions. Your task is to predict the next action of the last sequence based on the patterns observed in the provided input. Limit yourself to only answer with the predicted sequence, and follow the same format given as input.
\end{systemframe}
\end{samepage}
\vspace{0.5cm}

This prompt consistently offers essential context about the task, improving the model's performance, particularly in precision.

In Zero-Shot (ZS) settings, the instruction prompt is presented as:
\vspace{0.5cm}
\begin{samepage}
\begin{systemframe}
    \textit{System: } Below is an instruction that describes the task, paired with an input. Write a response that appropriately completes the request.
\end{systemframe}
\end{samepage}

Few-Shot prompting enhances this by adding examples of input sequences and corresponding responses to guide the model.

We explore two different input representations: a textual and a numerical one. In the first, we represent an action by its name and natural language, e.g., \texttt{attach-wheel}; in the second one, we use the index associated with the action triplet. 

\vspace{0.5cm}

\begin{samepage}
\begin{contextframe}
    \textbf{Textual Representation:}\\
    \textbf{Toy:} ``bulldozer"\\
    \textbf{Input Sequence:}\\
    attach-cabin, attach-body, attach-track, attach-blade\\
    \textbf{Next Symbol:}\\
    attach-figurine
\end{contextframe}
\end{samepage}

\vspace{0.5cm}
\noindent
\begin{samepage}
\begin{contextframe}
    \textbf{Numerical Representation:}\\
    \textbf{Toy:} ``bulldozer"\\
    \textbf{Input Sequence:}\\
    5678, 91011, 1213, 1415\\
    \textbf{Next Symbol:}\\
    1617
\end{contextframe}
\end{samepage}
\vspace{0.5cm}

In the \textbf{textual modality}, where actions are represented by their names (e.g., \texttt{attach-wheel}), adding the system prompt increases \(\mathrm{\text{Precision}}_{b}\) from 54.9\% to 56.7\% (+1.8\%), and Recall also rises from 86.5\% to 89.6\% (+3.1\%) , yielding an \(\mathrm{\text{F1 score}}_{b}\) gain from 67.1\% to 69.4\% (+2.3\%).

In the \textbf{numerical modality}, where actions are encoded by their index, including the system prompt boosts \(\mathrm{\text{Precision}}_{b}\) from 52.4\% to 54.5\% (+2.1\%) and Recall shifts from 94.6\% down to 90.3\% (–4.3\%), with a modest \(\mathrm{\text{F1 score}}_{b}\) decrease from 65.7\% to 67.9\% (+2.2\%).
The improvements obtained while using the textual representation can be attributed to the system prompt and few-shot examples, which provide explicit context and structural guidance, enabling the model to better understand the task and produce more accurate, relevant predictions.

\begin{table}[!t]
\caption{Results of LLAMA 3.1 8B on \assemblyO\ adding a system prompt (with corrected Precision and Recall values).}
\begin{tabular*}{\textwidth}{@{\extracolsep\fill}lcccc}
\toprule
System Prompt & Modality  & Precision\textsubscript{b} & Recall & F1 score\textsubscript{b} \\
\midrule
\multirow{2}{*}{\checkmark} 
              & \textbf{Textual}    & \textbf{56.7}     & 89.6   & \textbf{69.4}\\
              & Numerical  & 54.5      & 90.3   & 67.9 \\
\midrule
\multirow{2}{*}{X} 
              & Textual    & 54.9         & 86.5   & 67.1  \\
              & Numerical  & 52.4         & \textbf{94.6}   & 65.7  \\
\bottomrule
\end{tabular*}
\label{tab:system_prompt}
\end{table}

\paragraph{Prompting Methods.}
When comparing ZS, FS, and ACoT prompting methods, ACoT outperforms the rest of the prompting techniques (Table~\ref{table:prompt_analysis}), particularly when using the pre-trained model (\textit{Base}), across both textual and numerical tasks. 
In the ACoT approach, automatic ACoT reasoning is employed, where the LLM is queried twice: first, to generate an internal reasoning step for the answer, and then to use this reasoning, combined with a few examples, as in FS, to derive the final answer.
The first query is:

\vspace{0.5cm}
\begin{samepage}
\begin{acotframe}
Let's analyze each action in detail:
\begin{enumerate}
    \item For the action ``attach the cabin" consider the immediate effect and long-term consequences. 
    \item Repeat this analysis for each action in the sequence.
    \item Identify patterns and causal relationships between the actions.
\end{enumerate}
Now, let's proceed with the analysis step-by-step:
\end{acotframe}
\end{samepage}
\vspace{0.5mm}

This two-step process allows the model to provide an explanation for its decision and leverage the reasoning and examples to produce a more accurate and robust output.
In terms of the F1 score, the ACoT base (74.7\%) outperforms the ZS base (61.30\%) by 19.7\% and the FS base (69.50\%) by 7.2\%. 

These differences can be attributed to how these methods prompt the model. ZS and FS methods direct the model to generate answers straight from the input without guiding it through intermediate reasoning. Although this may suffice for more straightforward tasks, it often falls short when dealing with more complex tasks that require deeper understanding.

\begin{figure}[H]
    \includegraphics[width=1\textwidth]{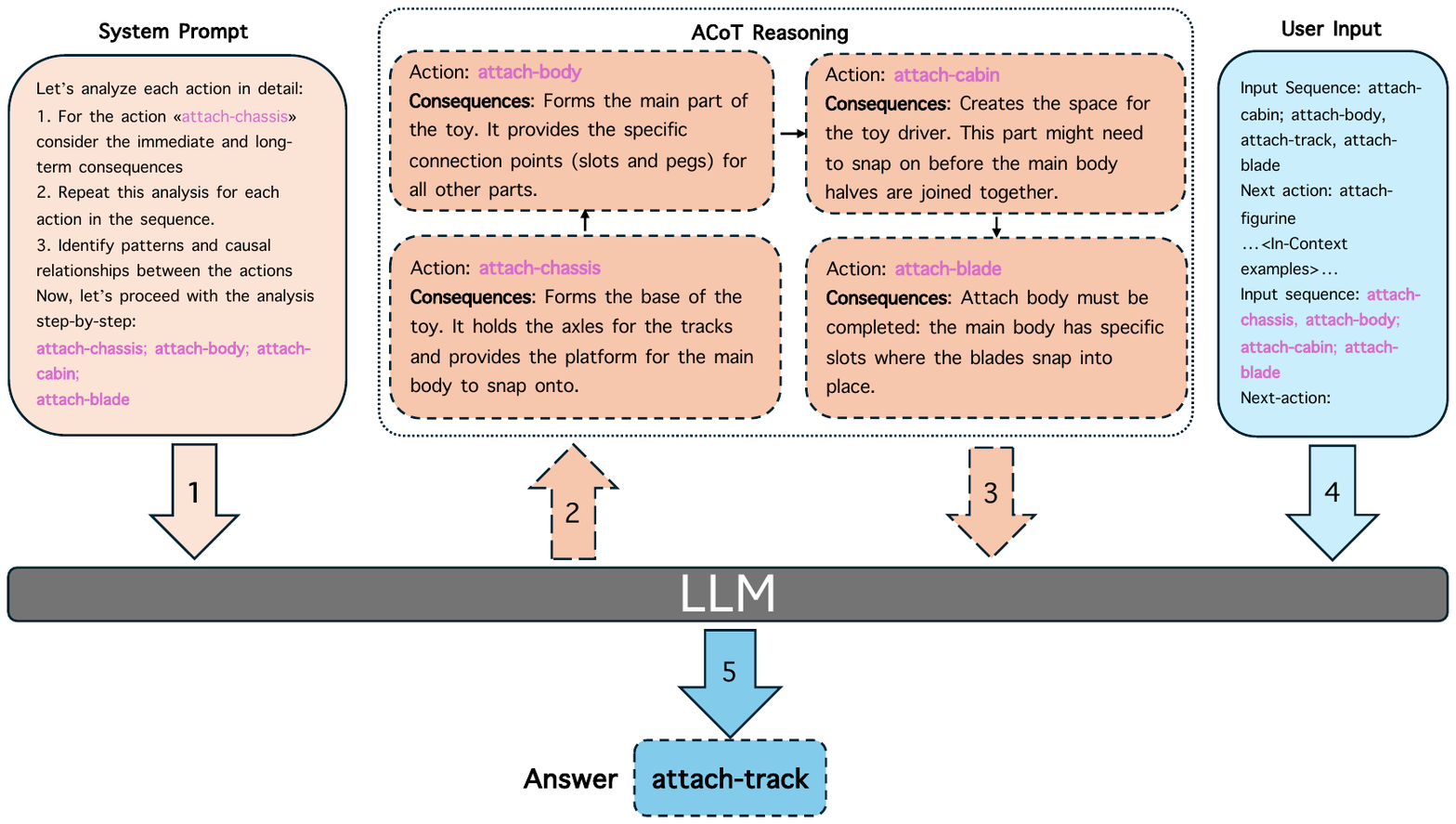}
    \caption{\added{Visualization of aCoT. First, the aCoT prompt is provided (1) to the LLM to enable (2) the generation of the CoT. The obtained reasoning is then fed to the model again together with the In-Context prompt (3, 4) to retrieve (5) the educated answer from the model.}}
    \label{fig: ACoT}
\end{figure}

ACoT, in contrast, excels in these scenarios because it prompts the model to break down the task into a series of logical intermediate steps before concluding \added{(Fig. \ref{fig: ACoT}).}
This approach mirrors the ``self-explanation'' cognitive strategy, where breaking down and explaining each step improves problem-solving skills. This explains why ACoT significantly outperforms ZS and FS, especially in tasks that demand nuanced understanding and intermediate reasoning.

\begin{table}[!t]
\caption{Results of LLAMA 3.1 8B on Assembly-101-O with different prompt methods, modalities, and inference modes.}\label{table:prompt_analysis}
\centering
\begin{tabular*}{\textwidth}{@{\extracolsep\fill}lccccc}
\toprule
Prompt & Modality & Mode & Precision\textsubscript{b} & Recall & F1 score\textsubscript{b} \\
\midrule
\multirow{4}{*}{ZS} 
    & Textual   & Base       & 47.1   & 87.9  & 61.3   \\
    & Textual   & Finetuned   & 48.4   & 93.4           & 63.8   \\
    & Numerical & Base        & 49.1   &     93.4       & 64.4   \\
    & Numerical & Finetuned   & 48.4   & 91.2           & 63.2   \\
\midrule
\multirow{4}{*}{FS} 
    & Textual   & Base       & 56.8   & 89.6           & 69.5   \\
    & Textual   & Finetuned   & 53.1   & 94.5           & 68.0   \\
    & Numerical & Base      & 54.5   & 90.1           & 67.9   \\
    & Numerical & Finetuned   & 55.4   & 86.9           & 67.6   \\
\midrule
\multirow{4}{*}{ACoT} 
    & Textual   & Base      &  \textbf{60.4}  & \textbf{97.8}           & \textbf{74.7}  \\
    & Textual   & Finetuned  & 56.3   & 87.9          & 68.6   \\
    & Numerical & Base        & 50.5   & 83.0           & 62.8   \\
    & Numerical & Finetuned   & 51.0   & 72.5           & 59.9   \\
\botrule
\end{tabular*}
\end{table}

\subsubsection{Finetuning Analysis}
\label{exp:finetuning}

We analyze the performance of LLAMA 3.1 8B when fine-tuned on \assemblyO (Table~\ref{table:prompt_analysis}). Contrary to the common expectation that fine-tuning enhances performance by adapting the model to the task's specific characteristics and distribution, this is not consistently observed. 
In both the ZS and FS methods, fine-tuning offers no significant improvement across metrics, with some cases showing similar or even lower \(\mathrm{\text{Precision}}_{b}\) and Recall values.
Notably, in the ACoT method, fine-tuning consistently results in a decrease in performance, particularly in terms of \
\(\mathrm{\text{Precision}}_{b}\), suggesting that fine-tuning may not always provide the anticipated benefits.

This phenomenon can be explained by considering the dataset's nature and ACoT prompting's capabilities. Fine-tuning typically helps when there is sufficient data to prevent overfitting, allowing the model to generalize better. However, when the dataset is small, as in this scenario, fine-tuning may cause the model to overfit the specific examples seen during fine-tuning, capturing noise rather than functional general patterns.

For ACoT, the base model's ability to handle complex reasoning through its intermediate steps seems sufficiently robust, and it does not benefit from the additional adjustments that fine-tuning provides. Fine-tuning might disturb these established reasoning pathways, leading to a decline in performance. This suggests that ACoT is inherently well-suited to tasks requiring deep reasoning, making it less dependent on the benefits of fine-tuning, especially in scenarios with limited data.

\subsubsection{Speed Analysis}
\label{exp:speed_analysis}

Table~\ref{table:speed_analysis} presents a speed test comparison of different prompting methods—ZS, FS, ACoT—using LLAMA 3.1 on the \assemblyO dataset. The results indicate that ZS is the fastest method, with an average processing speed of 0.208 seconds per sample. FS is slightly slower at 0.216 seconds per sample, representing a marginal 3.8\% rise in processing time compared to ZS. ACoT, while offering superior accuracy and reasoning ability, is the slowest, taking 0.315 seconds per sample, which is a 51.4\% increase in speed compared to ZS.

\begin{table}[!ht]
\caption{Speed Test with LLama 3.1 on Assembly-101-O}\label{tab2}
\begin{tabular*}{0.4\textwidth}{@{\extracolsep\fill}lc}
\toprule%
Model & Speed (sample/s) \\
\midrule
ZS              & \textbf{0.208} \\
FS                      & 0.216  \\
ACoT                         & 0.315 \\

\botrule
\end{tabular*}
\label{table:speed_analysis}
\end{table}

These differences in speed can be attributed to the inherent complexity of each method. ZS and FS prompt the model to generate answers directly, with FS introducing a minimal overhead due to the additional context provided by the few examples. On the other hand, ACoT requires the model to engage in more complex, step-by-step reasoning, which naturally demands more computational resources and time. While ACoT's performance benefits are clear, these come at the cost of slower processing, which could be a consideration in time-sensitive applications. Thus, the choice of prompting method should balance the trade-off between speed and output quality based on the task's specific requirements.

\added{For completeness, we also report the speed analysis of the language models used in Table~\ref{tab:principale}.  
The aggregation window is set to \( W = 200 \) frames at 30~fps, corresponding to an effective delay of approximately \( 200/30 \approx 6.67~\text{s} \).  
For all LLMs (See Table~\ref{tab:llm_speed}), inference latency is negligible by comparison.  
Thus, when decisions are made after temporal aggregation, the dominant latency stems from the windowing process required to stabilize per-frame recognition, rather than from the LLM inference itself.}

\begin{table}[!htp]
\caption{Speed test on baseline LLM methods.}
\label{tab:llm_speed}
\begin{tabular*}{0.9\textwidth}{@{\extracolsep\fill}lcc}
\toprule
Model & Anticipator speed (samples/s) & Total speed (steps/s) \\
\midrule
PREGO~\cite{Flaborea_2024_CVPR} -- GPT-3.5    & 1.436 & 8.096 \\
PREGO~\cite{Flaborea_2024_CVPR} -- DeepSeek-R1 & 0.612 & 7.272 \\
PREGO~\cite{Flaborea_2024_CVPR} -- LLaMA~3.1   & 0.216 & 6.876 \\
\textit{TI-PREGO} -- LLaMA~3.1                 & 0.315 & 6.975 \\
\bottomrule
\end{tabular*}
\end{table}

\subsection{Aggregation Strategy}
\label{sub:aggr}

\begin{figure}[H]
    \centering
    \includegraphics[width=1\linewidth]{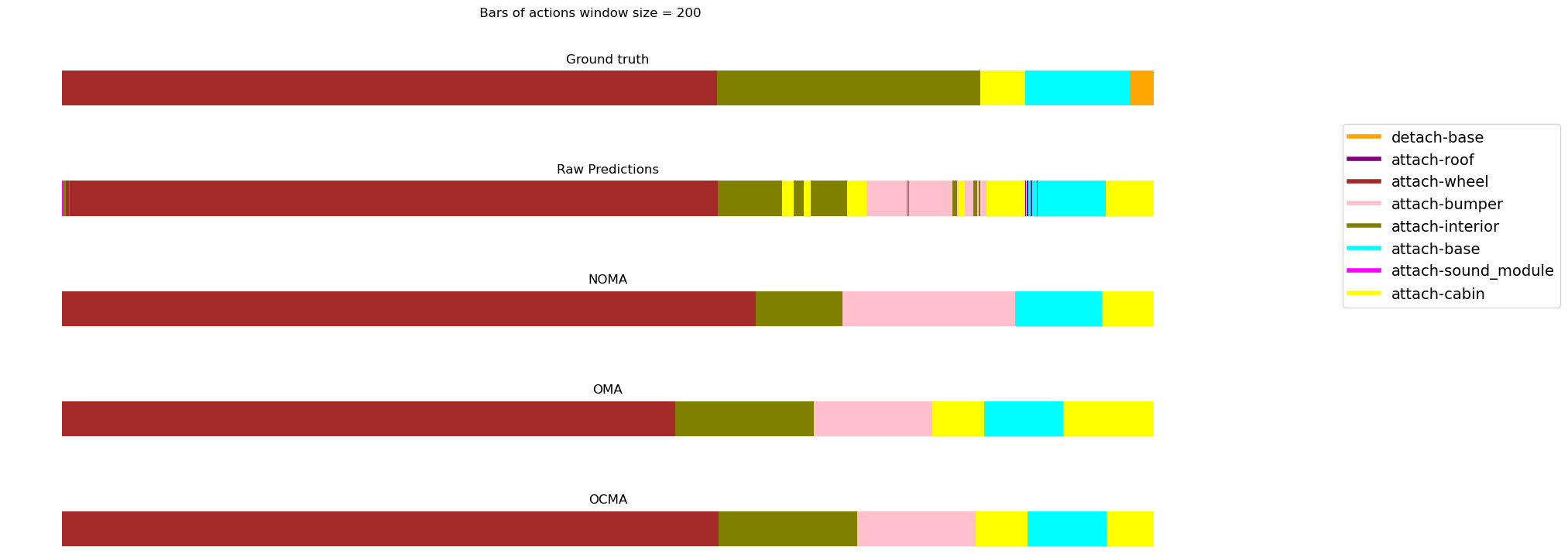}
    \caption{Predictions bars: (a) Ground truth predictions, (b) Predictions using the recognizer,(c) NOMA: Non-Overlapping Mode Aggregation, (d) OMA: Overlapping Mode Aggregation, (e) OCMA: Overlapping Centered Mode Aggregation }
    \label{fig:predictions bars}
\end{figure}

\added{In this section, we quantitatively evaluate the impact of each proposed aggregation strategy (cf. Sec.\ref{sec3:miniroad})}.
\added{As can be seen in Fig.\ref{fig:predictions bars}, all the three aggregation schemes}  
result
in a cleaner input for the anticipator module. 

\added{We assess the impact of the proposed aggregation strategies}
using the Levenshtein Similarity between the predicted and ground truth action sequence, which derives from the Levenshtein distance.
\added{With this choice,} 
the similarity between two action sequences (i.e. procedures) 
A and B is defined as:
\begin{equation}
    \text{Levenshtein Similarity}(A, B) = 1- \frac{d_{Lev}(A,B)}{\max(len(A), len(B))}
\end{equation}

where: $d_{Lev}(A,B)$ is the Levenshtein distance between A and 
B, which is the minimum number of single-action insertions, deletions or substitutions required to transform procedure A into procedure B, and $len(A)$ and $len(B)$ represent the length of procedure A and B, respectively. Note that the term $max(len(A),len(B))$ normalizes the score by the length of the longer procedure, ensuring that the ratio is between 0 and 1.

\begin{figure}[H]
    \centering
    \includegraphics[width=1\linewidth]{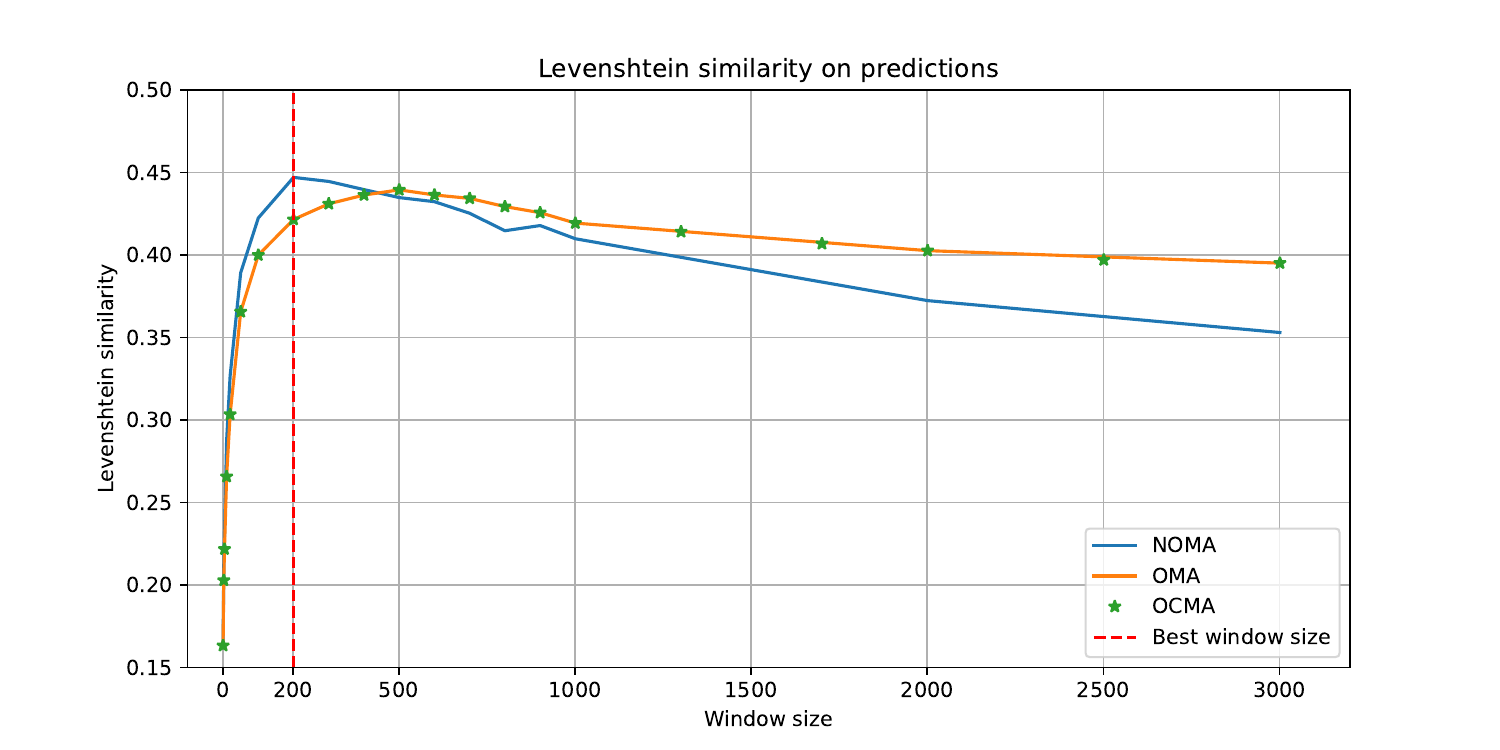}
    \caption{Levenshtein similarity for the proposed aggregation strategies. The maximum similarity is achieved by the first strategy when using a window size of 200 frames.}
    \label{fig:Levenshtein similarity}
\end{figure}

\begin{figure}[H]
    \centering
    \includegraphics[width=1\linewidth]{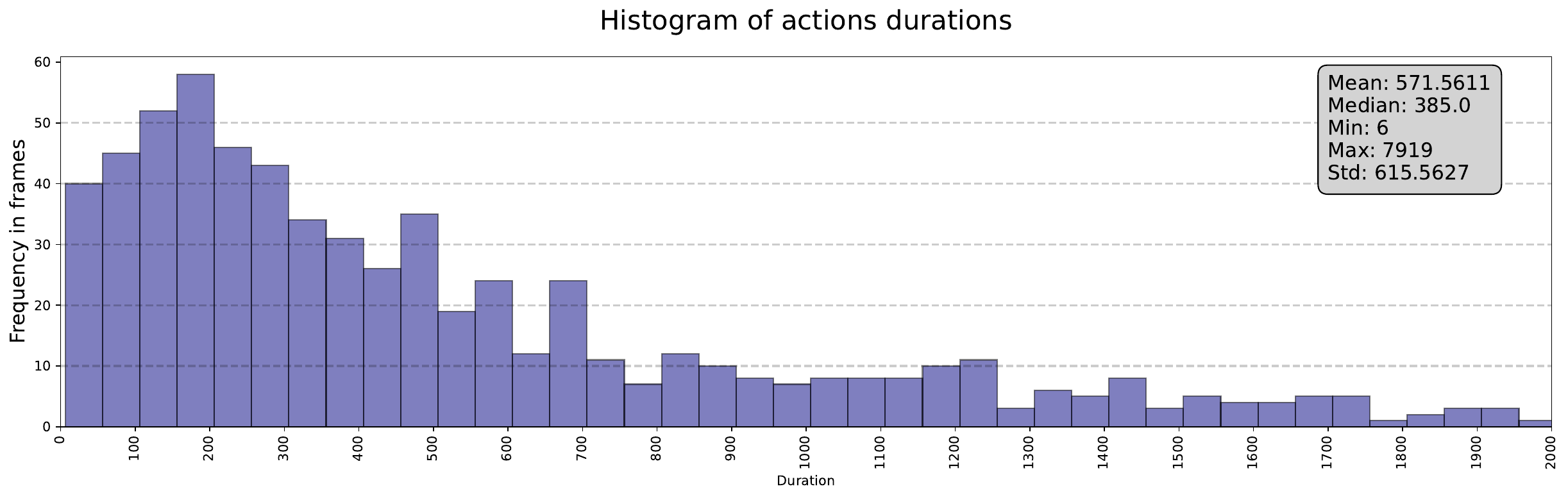}
    \caption{Histogram and statistics of the durations of the action in \assemblyO}
    \label{fig:Histogram Durations}
\end{figure}

\vspace{0.3cm}
Figure \ref{fig:Levenshtein similarity} compares the aggregated sequences with the respective ground truth using the Levenstain similarity. 
This highlights how the NOMA approach is the most effective in maximizing similarity among the proposed ones. 
Considering that datasets like \assembly are recorded at 30 fps and the average action lasts about 570 frames (Fig. \ref{fig:Histogram Durations}), we selected 200 frames as a sensible trade-off between achieving high similarity and minimizing the computational burden.

\section{Conclusion}\label{sec:conclusions}

This paper explored the challenging task of detecting procedural errors in egocentric videos online, focusing on a dual-branch architecture that combined action recognition and anticipation. By leveraging Large Language Models (LLMs) within the action anticipation branch, we used in-context learning and Automatic-Chain-of-Thought to demonstrate their ability to predict future steps based on prior actions. Our experiments comprehensively analyzed various prompting schemes and frame aggregation strategies to optimize performance in online action recognition, highlighting the importance of effective prompt formulation for LLM-based anticipators.

Our results emphasized the challenges of per-frame evaluation in real-time action recognition systems and showed how symbolic reasoning, combined with predictive models, enhanced procedural mistake detection.

Through extensive experimentation, we demonstrated the effectiveness of our dual-branch architecture, achieving new state-of-the-art performance for online mistake detection. This underscored the potential of integrating action recognition and anticipation in a unified framework for detecting procedural errors as they unfolded online. Future work could explore further refinements in symbolic representation and expand the range of procedural tasks and environments in which our framework can be applied.

\paragraph{Limitations}
\label{sec:limitations}
Our dual-branch architecture for online procedural mistake detection demonstrates promising results. In this section, we acknowledge some of its limitations.

Real-time processing constraints can present challenges, particularly when incorporating Large Language Models (LLMs), which may introduce latency during inference.
\added{To clarify the timing budget, our aggregation window of $W = 200$ frames at 30 fps induces an effective delay of approximately 6.7s, while the LLM anticipator adds on average $\sim0.3$,s per query. Hence, the dominant latency arises from the aggregation needed to stabilize per-frame recognition rather than from the LLM inference itself.}
Our focus is maintaining accuracy in an online setting, rather than optimizing for real-time performance.
\added{Nonetheless, as recognition quality improves, smaller windows may suffice, reducing aggregation-induced delay and making real-time optimization a promising direction for future work.}

While symbolic reasoning simplifies certain aspects of action prediction, it may not fully capture the complex semantic relationships between actions.
\added{In fact, LLMs, when used as in-context learners, can better capture such dependencies through contextual understanding and broad world knowledge.
However, their reasoning process is not explicitly based on logic rules and relies on patterns learned from data, which may limit their ability to model fine-grained or domain-specific action dependencies.}
Future research could explore hybrid approaches that integrate symbolic reasoning with richer representations.

\added{Our robustness treatment is pragmatic: aggregation reduces short-horizon fluctuations so that downstream anticipation operates on a steadier signal. Ambiguous cases, like brief transitions or rapidly alternating labels, are mitigated but not eliminated by this design. 
We use ACoT's intermediate explanations as qualitative aids to inspect typical successes and failures. These examples improve transparency but do not constitute a formal faithfulness guarantee. A systematic taxonomy of ambiguity and formal uncertainty quantification are valuable directions for future work and lie beyond the scope of this manuscript.}

\section*{Acknowledgements}

\noindent This work was carried out while Leonardo Plini was enrolled in the Italian National Doctorate on Artificial Intelligence run by Sapienza University of Rome. We thank DsTech S.r.l. and the PNRR MUR project PE0000013 Future Artificial Intelligence Research (FAIR) (CUP: B53C22003980006 and CUP: E63C22001940006) for partially funding the Sapienza University of Rome and University of Catania.

\section*{Data Availability Statement}

\noindent The datasets generated or analysed during the current study are available in (1) \href{https://sites.google.com/view/epic-tent}{https://sites.google.com/view/epic-tent} and (2) 
\href{https://assembly-101.github.io/}{https://assembly-101.github.io/}

\section*{AI Assistance Declaration}

During the preparation of this work, the authors used Large Language Models to assist in formulating the text.
After using these tools, the authors carefully reviewed and edited the content as necessary and take full responsibility for the final version of the published article.

\bibliography{sn-bibliography}

\end{document}